\documentclass[11pt]{article}

\usepackage[margin=1in]{geometry}
\usepackage{lmodern}
\usepackage{microtype}
\usepackage{amsmath,amssymb,amsthm,mathtools,bm}
\usepackage{booktabs}
\usepackage{natbib}
\usepackage{graphicx}
\usepackage{hyperref}
\usepackage{enumitem}
\usepackage{algorithm}
\usepackage{algpseudocode}
\usepackage{authblk}

\hypersetup{
    colorlinks=true
}

\title{Bayesian--AI Fusion for Epidemiological Decision Making: Calibrated Risk, Honest Uncertainty, and Hyperparameter Intelligence}
\author[1,2]{\textbf{Debashis Chatterjee}\thanks{Corresponding author. Email: \texttt{debashis.chatterjee@visva-bharati.ac.in}}}

\affil[1]{Department of Statistics, Visva--Bharati University, Santiniketan, India}
\affil[2]{ S.~N.~Bose National Centre for Basic Sciences, Kolkata, India}

\begin{document}
\maketitle



\begin{abstract}
Modern epidemiological analytics increasingly rely on machine learning models that deliver accurate predictions but often fail to quantify uncertainty or support transparent decision rules. In parallel, Bayesian methods offer principled uncertainty quantification and calibration, yet are frequently perceived as computationally expensive and difficult to integrate with contemporary AI pipelines. This paper proposes a unified \emph{Bayesian--AI framework} that explicitly fuses these two paradigms.

On the predictive side, we use Bayesian logistic regression to obtain calibrated individual-level probabilities and credible intervals for binary disease outcomes, illustrated on the Pima Indians Diabetes cohort. On the algorithmic side, we use Gaussian-process Bayesian Optimization to navigate noisy, expensive hyperparameter landscapes in penalised Cox survival models, demonstrated on the \texttt{GBSG2} breast cancer dataset. A two-layer architecture emerges: (i) a Bayesian predictive layer that treats risk as a posterior distribution rather than a point estimate, and (ii) a Bayesian optimisation layer that treats model selection and hyperparameter tuning as inference over a black-box objective.

To verify that the proposed framework behaves in a statistically honest way, we conduct simulation studies under known ground truth, including a deliberately challenging high-dimensional, small-sample, correlated-covariate regime. In well-specified low-dimensional settings, Bayesian and frequentist estimators achieve similar discrimination, but the Bayesian layer additionally delivers calibrated credible intervals and coverage. In high-dimensional regimes, Bayesian shrinkage yields strictly better AUC, Brier score, log-loss, and calibration slopes than unregularised maximum likelihood, while Bayesian Optimization systematically pushes penalised Cox models towards near-oracle concordance.

Taken together, these results support a broader methodological thesis: Bayesian reasoning should inform both \emph{what} we infer (posterior predictive risk) and \emph{how} we search (hyperparameters and model classes), providing calibrated risk, honest uncertainty, and principled hyperparameter intelligence for epidemiological decision making.
\end{abstract}

\bigskip

\noindent\textbf{Keywords:}
Bayesian inference; uncertainty quantification; Bayesian Optimization; epidemiology; survival analysis; calibration; hyperparameter tuning; probabilistic machine learning.

\section{Introduction}
\label{sec:intro}

Epidemiological decision making involves uncertainty at every stage: uncertainty about disease status, uncertainty about future clinical progression, and uncertainty about which models are appropriate in the first place. Classical statistical models provide interpretable structures but are often rigid. Modern machine-learning models provide accuracy but lack interpretability and calibrated uncertainty. Bayesian methods provide coherent uncertainty quantification but often struggle with the increasing algorithmic complexity of modern AI pipelines.

This paper argues that a \emph{fusion} of Bayesian inference and AI principles provides a powerful and necessary methodological foundation for epidemiological prediction and decision making. The fusion operates on two layers:

\begin{enumerate}[leftmargin=1.2em]
\item \textbf{Layer 1 --- Bayesian Inference for Prediction:}  
Individual-level predictive probabilities are treated as full posterior distributions rather than point estimates. This resolves a central problem in clinical analytics: how to select decisions under uncertainty.

\item \textbf{Layer 2 --- Bayesian Optimization for Model Selection:}  
Hyperparameter tuning, model selection, and architecture search are themselves inference problems under uncertainty. Bayesian Optimization provides a principled way to explore these high-dimensional landscapes efficiently.
\end{enumerate}

The two layers operate at different levels of abstraction but share the same epistemic foundation. The first quantifies the uncertainty \emph{within} a predictive model; the second quantifies uncertainty \emph{over} a landscape of models.

\subsection{Motivation}
Accurate prediction is no longer sufficient for modern epidemiological practice. Screening, treatment, and follow-up policies must account for uncertainty, risk calibration, and the cost structure of false decisions. An AI system that produces good point predictions but poor uncertainty estimates can be harmful. Conversely, a Bayesian model with beautiful uncertainty quantification but poor tuned hyperparameters may underperform in practice.

The proposed Bayesian–AI fusion bridges this gap.

\subsection{Two Running Examples}
\begin{itemize}
\item \textbf{Binary outcome (diabetes)}:  
Use Bayesian logistic regression to model diabetes risk in the \texttt{PimaIndiansDiabetes} dataset. Derive posterior predictive probabilities and credible intervals for individual patients. Evaluate model calibration and discuss decision-theoretic thresholds.

\item \textbf{Survival outcome (breast cancer)}:  
Use Bayesian Optimization to tune a penalized Cox model for recurrence-time prediction in the \texttt{GBSG2} dataset. The optimization surface is noisy and expensive; Bayesian Optimization efficiently identifies high-performing hyperparameter regions.
\end{itemize}

These applications demonstrate the vertical integration of Bayesian tools into AI pipelines.

\section{Related Work}
\label{sec:related}

The proposed Bayesian--AI framework for epidemiological decision making sits at the intersection of
(i) Bayesian deep learning and uncertainty quantification,
(ii) Bayesian optimisation for hyperparameter tuning and AutoML,
(iii) statistical and AI-based epidemiological modelling, and
(iv) calibration and decision-focused evaluation of predictive systems.
In what follows we situate our contribution within these strands of literature and highlight the gap that motivates our framework.

\subsection{Bayesian Deep Learning and Predictive Uncertainty}

Bayesian views of neural networks date back at least to \citet{Neal1996}, who framed multilayer perceptrons as priors over functions and used Markov chain Monte Carlo for posterior inference.
This line of work underpins modern Bayesian deep learning, where approximate inference is required for scalability.
Variational approaches such as Bayes-by-Backprop \citep{Blundell2015} and Monte Carlo dropout \citep{Gal2016} provide tractable approximations to the weight posterior,
while \citet{Kendall2017} distinguish between aleatoric and epistemic uncertainties and show how to integrate both in computer vision tasks.
Deep ensembles \citep{Lakshminarayanan2017} offer a strong practical baseline for predictive uncertainty without explicit Bayesian priors,
and have become widely adopted in safety-critical settings.

At a broader level, \citet{Ghahramani2015} and \citet{Blei2017} argue that probabilistic machine learning is a natural language for reasoning under uncertainty in AI systems.
Recent empirical studies critically examine how well Bayesian posteriors behave in deep networks:
\citet{Wenzel2020} show that na\"{i}ve posterior sampling can still yield miscalibrated predictions,
while \citet{Ovadia2019} and \citet{Abdar2021} systematically evaluate uncertainty quality under distribution shift.
Our use of Bayesian logistic regression with full posterior predictive distributions is conceptually simpler than deep architectures,
but is aligned with this literature in treating uncertainty as a first-class object.

\subsection{Bayesian Optimisation and Automated Model Tuning}

Hyperparameter tuning and architecture search have long been recognised as key bottlenecks in applied machine learning.
Random search \citep{Bergstra2012} improves substantially over hand-designed grids,
but treats evaluations as independent.
Bayesian optimisation \citep{Snoek2012,Frazier2018} instead places a Gaussian process prior on the unknown performance surface and uses acquisition functions to balance exploration and exploitation.
These ideas have been further developed in the AutoML community,
e.g., in \citet{Feurer2015} and the survey book of \citet{Hutter2019},
where Bayesian optimisation is combined with meta-learning and pipeline search.

Closer to our application, \citet{Alaa2018} introduced AutoPrognosis, a Bayesian-optimisation-based AutoML framework for clinical risk prediction that tunes ensembles of survival and classification models,
and \citet{Imrie2023} extended this to AutoPrognosis~2.0 with a focus on fairness and interpretability in healthcare.
Our use of Gaussian-process-based Bayesian optimisation to tune penalised Cox models is inspired by these works,
but specialised to a low-dimensional hyperparameter space $(\log\lambda,\alpha)$ and explicitly framed as a layer of \emph{Bayesian meta-inference} on top of classical survival models.

\subsection{Bayesian and AI-Driven Epidemiological Modelling}

There is a long tradition of Bayesian statistical inference for mechanistic infectious disease models.
\citet{Toni2009} developed approximate Bayesian computation schemes for parameter inference and model selection in dynamical systems,
and \citet{Kypraios2017} surveyed Bayesian methods for transmission models using household data.
\citet{Minter2019} reviewed approximate Bayesian computation for infectious-disease modelling,
highlighting the tension between realistic mechanistic models and computational tractability.

More recently, emulator-based methods and Bayesian optimisation have been proposed to accelerate calibration of complex transmission models.
For example, \citet{Reiker2021} use Gaussian-process emulators and Bayesian optimisation to design malaria intervention policies under uncertainty,
illustrating that optimising over control strategies can itself be seen as a Bayesian decision problem.
Beyond purely statistical or purely mechanistic models,
\citet{Ye2025} provide a comprehensive scoping review of integrated AI--mechanistic approaches for epidemiological modelling across a wide range of pathogens,
and \citet{Kraemer2024} survey AI applications for infectious-disease prediction, surveillance and control.

Parallel developments in mathematical epidemiology \citep{Brauer2008} and multi-model forecasting \citep{Reich2019} have emphasised the value of ensembles, probabilistic forecasts and decision-relevant metrics for public health planning.
Work on the interpretation and limitations of epidemic forecasts, such as \citet{Holmdahl2020}, has underscored the need for honest uncertainty quantification and transparent model assumptions.
Our framework aligns with this literature by treating both binary risk prediction and survival forecasting as probabilistic tasks, while using Bayesian optimisation to navigate complex model spaces in a computationally efficient way.

\subsection{Machine Learning for Risk Prediction and Survival in Health}

Machine-learning-based risk prediction has become central in clinical decision support.
\citet{Miotto2016} introduced Deep Patient, an unsupervised representation-learning approach on electronic health records (EHRs),
while \citet{Rajkomar2018} demonstrated scalable deep learning models for a variety of EHR prediction tasks.
In primary care, \citet{Weng2017} compared traditional and machine-learning models for cardiovascular risk prediction,
showing that modern ML can improve discrimination but raising questions about calibration and interpretability.
In parallel, \citet{Reich2019} coordinated multi-model influenza forecasting efforts, providing a template for rigorous out-of-sample evaluation of predictive systems in epidemiology.

Our work differs from many of these studies in two respects.
First, we keep the base models comparatively simple (logistic regression and elastic-net Cox regression),
and focus instead on the \emph{Bayesian wrapper}: posterior predictive risk, credible intervals, and Bayesian optimisation.
Second, we explicitly link these modelling choices to decision-theoretic quantities such as cost-weighted screening thresholds and concordance-driven hyperparameter selection,
placing the emphasis on calibrated, uncertainty-aware decision support rather than maximal predictive accuracy alone.

\subsection{Calibration, Reliability and Decision-Focused Evaluation}

A key theme across modern ML for health is that good point predictions are not sufficient for safe deployment.
Calibration metrics and reliability diagrams are now standard tools for assessing probabilistic forecasts \citep{Gneiting2007,Guo2017}.
Classical approaches to calibration, such as Platt scaling \citep{Platt1999} and isotonic regression \citep{NiculescuMizil2005},
remain widely used to post-process scores from black-box models.
\citet{Begoli2019} argue that uncertainty quantification is essential for machine-assisted medical decision making,
and \citet{Abdar2021} provide an extensive review of uncertainty quantification techniques in deep learning.

In the epidemiological context, \citet{Holmdahl2020} document how overconfident or poorly calibrated forecasts can mislead policy.
Similarly, \citet{Ovadia2019} and \citet{Wenzel2020} show that distribution shifts and approximate posteriors can degrade uncertainty estimates in deep networks.
Our framework responds to these concerns in two complementary ways:
(i) by using fully Bayesian predictive distributions in the binary risk example, enabling direct assessment of calibration and interval coverage;
and (ii) by using Bayesian optimisation to treat model performance as a random function over hyperparameters, explicitly modelling the uncertainty in which model configuration should be deployed.

\subsection{Summary and Gap}

Taken together, the existing literature provides powerful building blocks:
Bayesian deep learning for uncertainty, Bayesian optimisation for model search, mechanistic and data-driven epidemic models, and calibration-focused evaluation.
However, most works address these components in isolation.
AutoML frameworks such as AutoPrognosis \citep{Alaa2018,Imrie2023} integrate Bayesian optimisation with clinical prediction models,
but do not explicitly foreground calibration and decision-theoretic interpretation.
Conversely, epidemiological forecasting studies \citep{Reich2019,Ye2025,Kraemer2024} emphasise probabilistic forecasts and public health relevance,
but rarely treat hyperparameter tuning itself as a Bayesian inference problem.

Our contribution is to synthesise these strands into a \emph{two-layer} Bayesian--AI architecture for epidemiological decision making:
a Bayesian predictive layer that yields calibrated individual-level risk and uncertainty,
and a Bayesian optimisation layer that treats model selection and tuning as probabilistic inference over hyperparameters.
The simulation and real-data studies in Sections~\ref{sec:results} and~\ref{sec:simulation} show that this fusion is not merely conceptually appealing, but also empirically competitive in terms of discrimination, calibration and concordance.

\section{Research Objectives and Methodological Novelty}
\label{sec:objectives}

This paper has three core research objectives:

\begin{enumerate}[label=\textbf{O\arabic*},leftmargin=1.2em]
\item \textbf{To construct calibrated Bayesian predictive models for epidemiological risk analysis.}  
We aim to obtain individual-level posterior predictive distributions, not merely point estimates, enabling risk stratification and cost-based decision rules.

\item \textbf{To demonstrate Bayesian Optimization as a decision-theoretic tool for hyperparameter selection in survival models.}  
Survival models (e.g., penalized Cox models) often require tuning of penalty strength and sparsity levels. Bayesian Optimization efficiently explores such hyperparameter landscapes under uncertainty.

\item \textbf{To establish a unified Bayesian–AI conceptual framework for epidemiological modelling.}  
We articulate a view where Bayesian reasoning governs both inference \emph{and} model search.
\end{enumerate}

\subsection{Novelty}
The novelty of this work lies in synthesizing two areas usually treated independently:

\begin{itemize}
\item Bayesian inference of \emph{predictive probabilities} and posterior risk for binary outcomes.
\item Bayesian Optimization of \emph{hyperparameters} in survival modelling.
\end{itemize}

The conceptual leap is recognizing that hyperparameter search is itself a Bayesian decision problem. This reframes model selection as a process of \emph{active learning over the model space}, rather than as a mechanical grid search.


\section{Preliminaries}
\label{sec:preliminaries}

This section collects the basic probabilistic and methodological concepts underpinning
our Bayesian--AI framework. The goal is not to re-derive the full models of
Section~\ref{sec:methodology}, but to provide a coherent conceptual backbone for the
notation and constructions used later. Throughout, random quantities are denoted in
uppercase (e.g.\ $Y$), observed data in lowercase (e.g.\ $y$), and boldface symbols
(e.g.\ $\mathbf{y}$, $\mathbf{x}$) denote vectors or collections.

\subsection{Bayesian Inference and Predictive Distributions}
\label{subsec:prelim_bayes}

Let $\theta$ denote an unknown parameter (possibly vector-valued) and
$\mathcal{D}=\{(x_i,y_i)\}_{i=1}^n$ denote observed data. In the Bayesian paradigm, we
specify a prior $p(\theta)$, a likelihood $p(\mathcal{D}\mid\theta)$, and obtain the
posterior via Bayes' rule \citep{Gelman2013BDA,Berger1985}:
\begin{equation}
  p(\theta\mid\mathcal{D})
  =
  \frac{p(\mathcal{D}\mid\theta)\,p(\theta)}{p(\mathcal{D})},
  \qquad
  p(\mathcal{D})=
  \int p(\mathcal{D}\mid\theta)\,p(\theta)\,d\theta.
  \label{eq:bayes_rule_generic}
\end{equation}
In most realistic models, the evidence $p(\mathcal{D})$ is intractable, and we rely on
Markov chain Monte Carlo or related simulation-based methods to explore the posterior
\citep{Gelman2013BDA,Bishop2006,Blei2017VI}.

A central object in decision making is the \emph{posterior predictive distribution} for
a new outcome $Y^\ast$ at covariate $x^\ast$:
\begin{equation}
  p(y^\ast\mid x^\ast,\mathcal{D})
  =
  \int p(y^\ast\mid x^\ast,\theta)\,p(\theta\mid\mathcal{D})\,d\theta.
  \label{eq:post_pred_generic}
\end{equation}
In Section~\ref{subsec:bayes_pred}, the logistic model in \eqref{eq:logistic} and its
posterior \eqref{eq:priors}--(implicit) are special cases of this generic structure,
with $\theta=(\beta_0,\beta)$ and $p(y^\ast\mid x^\ast,\theta)$ given by the Bernoulli
likelihood induced by the logit link.

The Bayesian decision-theoretic screening rule in
\eqref{eq:bayes_threshold} is derived by combining the posterior predictive
\eqref{eq:post_pred_generic} with a loss function on actions and outcomes, following
classical treatments in \citet{Berger1985,Gelman2013BDA}.

\subsection{Generalized Linear Models and Logistic Regression}
\label{subsec:prelim_glm}

Generalized linear models (GLMs) provide a unifying framework for regression with
non-Gaussian responses \citep{Nelder1972GLM,Hastie2009ESL}. A GLM assumes (i) a
conditional distribution $Y_i\mid x_i,\theta$ in an exponential family, and
(ii) a link function $g$ relating the mean $\mu_i=\mathbb{E}[Y_i\mid x_i,\theta]$ to a
linear predictor $\eta_i$:
\[
  g(\mu_i) = \eta_i = x_i^\top\beta.
\]
The logistic regression model in \eqref{eq:logistic} is the canonical GLM for binary
responses, with the logit link
$g(\mu)=\log\{\mu/(1-\mu)\}$ and Bernoulli likelihood
\eqref{eq:logistic_lik}. The Gaussian priors in \eqref{eq:priors} yield a
\emph{Bayesian GLM} \citep{Bishop2006,Goodrich2025rstanarm}, whose posterior is
approximated via Hamiltonian Monte Carlo in \texttt{rstanarm}.

Within this GLM framework, discrimination metrics such as the ROC curve and AUC are
computed from predictive probabilities $\hat{p}_i =
\Pr(Y_i=1\mid x_i,\mathcal{D})$ as in \eqref{eq:post_pred_mc}, following standard
diagnostic-classification practice \citep{Pepe2003ROC}. Calibration, defined in terms
of agreement between $\hat{p}_i$ and empirical event frequencies, is assessed via
binning and plotting as in \citet{Guo2017Calibration}; this motivates the decile-based
calibration procedure described in Section~\ref{subsec:bayes_pred}.

\subsection{Survival Analysis and Concordance}
\label{subsec:prelim_survival}

Survival analysis concerns time-to-event data subject to censoring
\citep{Cox1972,Harrell2015RMS}. Each individual $i$ contributes a pair
$Y_i=(T_i,\delta_i)$, where $T_i$ is the observed time and $\delta_i\in\{0,1\}$ is an
event indicator (1 = event, 0 = right-censored). The \emph{survival function} is
$S(t)=\Pr(T>t)$ and the \emph{hazard function} is
$h(t)=\lim_{\Delta t\to 0^+}\Pr(t\le T<t+\Delta t\mid T\ge t)/\Delta t$.
In the Cox proportional hazards model \eqref{eq:cox_hazard}, we posit
\[
  h(t\mid X_i) = h_0(t)\exp(X_i^\top\beta),
\]
where $h_0(t)$ is an unspecified baseline hazard and $\beta$ are regression
coefficients estimated via the partial likelihood \citep{Cox1972}.

For model evaluation, we use the concordance index (C-index), which measures the
probability that, for a randomly chosen pair of comparable individuals, the subject
with the higher predicted risk (or linear predictor) experiences the event first
\citep{Harrell2015RMS}. In the penalised Cox setting of
Section~\ref{subsec:cox_bo}, the C-index is computed on a validation subset using the
risk scores $\ell_i(\lambda,\alpha)=X_i^\top\hat{\beta}(\lambda,\alpha)$, and serves
as the black-box objective $f(\theta)$ for Bayesian Optimization.

\subsection{Gaussian Processes and Bayesian Optimization}
\label{subsec:prelim_gp_bo}

Gaussian processes (GPs) are flexible priors over functions
$f:\Theta\to\mathbb{R}$ \citep{RasmussenWilliams2006GPML}. A GP is specified by a mean
function $m(\theta)$ and covariance kernel $k(\theta,\theta')$; for any finite set
$\{\theta_1,\dots,\theta_T\}$, the vector
$\bigl(f(\theta_1),\dots,f(\theta_T)\bigr)$ is multivariate Gaussian with mean
$\bigl(m(\theta_1),\dots,m(\theta_T)\bigr)$ and covariance matrix
$[k(\theta_s,\theta_t)]_{s,t}$. Conditioning on noisy observations
$\{(\theta_t,f_t)\}$ yields closed-form expressions for the posterior mean
$\mu_T(\theta)$ and variance $s_T^2(\theta)$ of $f(\theta)$
\citep{RasmussenWilliams2006GPML}.

Bayesian Optimization (BO) uses a GP surrogate to optimise an expensive black-box
function $f(\theta)$ \citep{Snoek2012PracticalBO,Frazier2018BO}. Given the GP posterior,
an \emph{acquisition function} $a_T(\theta)$ encodes the utility of evaluating
$f$ at $\theta$, balancing exploration (large $s_T(\theta)$) and exploitation (large
$\mu_T(\theta)$). The Upper Confidence Bound (UCB) acquisition in \eqref{eq:ucb} is a
standard choice:
\[
  a_T(\theta) = \mu_T(\theta) + \kappa\,s_T(\theta),
\]
with $\kappa>0$ controlling the exploration--exploitation trade-off. The next
evaluation point is chosen as $\theta_{T+1} = \arg\max_{\theta\in\Theta} a_T(\theta)$,
and the GP is updated after observing $f(\theta_{T+1})$. In
Section~\ref{subsec:cox_bo}, this machinery is applied to the hyperparameter vector
$\theta=(\log\lambda,\alpha)$ of the elastic-net penalised Cox model, with
$f(\theta)$ equal to the validation C-index.

This GP-based BO framework places a \emph{Bayesian} distribution over functions
representing model performance, thereby elevating hyperparameter tuning itself to an
inference problem. In combination with the Bayesian predictive modelling of
Section~\ref{subsec:bayes_pred}, this yields the two-layer Bayesian--AI architecture
that we exploit in our epidemiological applications.

\subsection{Evaluation Metrics for Probabilistic Predictions}
\label{subsec:prelim_metrics}

Finally, we briefly summarise the key evaluation metrics used throughout the paper.

\paragraph{Discrimination.}
For binary outcomes, the ROC curve and area under the curve (AUC) quantify the ability
of a model to rank positive cases above negative cases
\citep{Pepe2003ROC,Hastie2009ESL}. In our framework, these are computed on the test
set using posterior predictive means $\hat{p}_i$ from
\eqref{eq:post_pred_mc}. For survival outcomes, discrimination is measured by the
C-index as discussed above.

\paragraph{Calibration.}
Calibration assesses whether predicted probabilities agree with observed frequencies
\citep{Guo2017Calibration}. We use decile-binned calibration tables and plots, where
perfect calibration corresponds to points lying on the diagonal. Bayesian models with
well-chosen priors and likelihoods often exhibit good calibration properties
\citep{Gelman2013BDA}, but empirical checking remains essential in applied work.

These preliminaries provide the conceptual scaffolding for the specific models and
algorithms instantiated in Section~\ref{sec:methodology}.

\section{Methodology}
\label{sec:methodology}

Our methodological contribution is organised around a simple but powerful idea:
\emph{Bayesian thinking should inform both prediction and model search}. Concretely, we
combine (i) Bayesian predictive modelling for individual-level risk estimation in a
binary outcome setting and (ii) Bayesian Optimization for hyperparameter tuning in
penalised survival models. The first component focuses on uncertainty \emph{within} a
given model; the second focuses on uncertainty \emph{over} a space of models and
hyperparameters. Conceptually, this follows the broader paradigm of probabilistic
machine learning, where models, predictions, and even algorithmic choices are framed as
inference under uncertainty \citep{Ghahramani2015,Bishop2006}.

\subsection{Data and Preprocessing}
\label{subsec:data}

\subsubsection{Pima Indians Diabetes (Binary Outcome)}

For the binary risk-modelling component, we use the well-known
\texttt{PimaIndiansDiabetes} dataset distributed in the \texttt{mlbench} R package
\citep{Leisch2024mlbench}. The data contain $n = 768$ observations on female Pima
Indian patients, with the binary outcome
\[
  y_i \in \{0,1\}, \qquad y_i = 1 \; \text{indicating diagnosed diabetes},
\]
and covariate vector $x_i \in \mathbb{R}^p$ including number of pregnancies, plasma
glucose concentration, diastolic blood pressure, triceps skinfold thickness, two-hour
serum insulin, body-mass index (BMI), diabetes pedigree function, and age. Covariates
are used on their original scales except where otherwise noted; all modelling is
conducted in R, and the dataset used is the canonical \texttt{PimaIndiansDiabetes}
object provided by \citet{Leisch2024mlbench}.

We randomly partition the data into a training set $\mathcal{D}_{\mathrm{train}}$
($70\%$) and a test set $\mathcal{D}_{\mathrm{test}}$ ($30\%$), using a fixed random
seed for reproducibility. All model fitting and posterior sampling for the Bayesian
logistic regression is performed on $\mathcal{D}_{\mathrm{train}}$, while predictive
performance and calibration are evaluated on $\mathcal{D}_{\mathrm{test}}$.

\subsubsection{GBSG2 Breast Cancer Survival (Time-to-Event Outcome)}

For the survival-modelling component, we use the \texttt{GBSG2} dataset from the
\texttt{TH.data} R package \citep{Hothorn2019THdata}, which reports recurrence-free
survival times for $n = 686$ women enrolled in the German Breast Cancer Study Group~2
trial. The dataset includes hormonal therapy indicator, age, menopausal status, tumour
size and grade, number of positive nodes, progesterone receptor, and oestrogen
receptor measurements, together with event time $T_i$ and censoring indicator
$\delta_i \in \{0,1\}$.

We form the standard survival response
\[
  Y_i = (T_i, \delta_i), \qquad i=1,\dots,n,
\]
and create design matrices for penalised Cox regression using \texttt{model.matrix} in
R. As with the diabetes case, we randomly split the data into training and validation
subsets (70/30). The training subset is used to fit penalised Cox models; the
validation subset is used to compute concordance indices for Bayesian Optimization of
hyperparameters.

\subsection{Bayesian Predictive Modelling for Binary Risk}
\label{subsec:bayes_pred}

\subsubsection{Model Specification}

Let $\mathcal{D} = \{(x_i, y_i): i = 1,\dots,n\}$ denote the training data from the
Pima cohort. We model the conditional probability of diabetes using a Bayesian
logistic regression model \citep{Bishop2006}:
\begin{equation}
  \Pr(Y_i = 1 \mid x_i, \beta)
  \;=\;
  \sigma(\eta_i),
  \qquad
  \eta_i = \beta_0 + x_i^\top \beta,
  \qquad
  \sigma(z) = \frac{1}{1 + e^{-z}},
  \label{eq:logistic}
\end{equation}
where $\beta_0 \in \mathbb{R}$ is an intercept and $\beta \in \mathbb{R}^p$ is the
coefficient vector. Conditional on $(\beta_0,\beta)$, the likelihood factorises as
\begin{equation}
  p(\mathbf{y} \mid \beta_0,\beta)
  \;=\;
  \prod_{i=1}^n
  \sigma(\eta_i)^{\,y_i}
  \bigl\{1 - \sigma(\eta_i)\bigr\}^{1-y_i}.
  \label{eq:logistic_lik}
\end{equation}

We adopt weakly informative Gaussian priors,
\begin{equation}
  \beta_0 \sim \mathcal{N}(0,\tau_0^2),
  \qquad
  \beta_j \sim \mathcal{N}(0,\tau^2),
  \quad j=1,\dots,p,
  \label{eq:priors}
\end{equation}
with scales $\tau_0,\tau$ chosen large enough to regularise extreme values without
unduly constraining the posterior, following general recommendations for Bayesian
generalised linear models \citep{Bishop2006,Goodrich2025rstanarm}.

\subsubsection{Posterior Inference via \texttt{rstanarm}}

Combining \eqref{eq:logistic_lik} and \eqref{eq:priors}, the posterior distribution
\[
  p(\beta_0,\beta \mid \mathcal{D})
  \;\propto\;
  p(\mathbf{y}\mid\beta_0,\beta)\,
  p(\beta_0)\,p(\beta)
\]
is analytically intractable, making numerical approximation necessary. We use the
\texttt{rstanarm} package \citep{Goodrich2025rstanarm}, which provides a
high-level interface to Stan’s Hamiltonian Monte Carlo algorithms for Bayesian
regression modelling. Specifically, we fit the model using \texttt{stan\_glm} with a
binomial logit link, running multiple Markov chains and checking convergence via the
potential scale reduction statistic $\hat{R}$ and effective sample size diagnostics.

While alternative approximate methods exist---including variational inference
\citep{Blei2017VI} and expectation propagation \citep{Bishop2006}---we prefer HMC
here to ensure accurate posterior uncertainty quantification, which is central to our
application.

\subsubsection{Posterior Predictive Distribution and Uncertainty}

For a new individual with covariates $x^\ast$, the posterior predictive probability of
diabetes is
\begin{equation}
  p\bigl(y^\ast = 1 \mid x^\ast, \mathcal{D}\bigr)
  \;=\;
  \int \sigma(\beta_0 + x^{\ast\top}\beta)\,
  p(\beta_0,\beta\mid\mathcal{D})\,
  d\beta_0\,d\beta.
  \label{eq:post_pred}
\end{equation}
Given $S$ posterior draws $\{(\beta_0^{(s)},\beta^{(s)})\}_{s=1}^S$, we approximate
\eqref{eq:post_pred} by
\begin{equation}
  \hat{p}^\ast
  \;=\;
  \frac{1}{S}\sum_{s=1}^S
  \sigma\!\bigl(\beta_0^{(s)} + x^{\ast\top}\beta^{(s)}\bigr),
  \label{eq:post_pred_mc}
\end{equation}
and obtain a $(1-\alpha)$ credible interval for $p^\ast$ from the empirical
$\alpha/2$ and $1-\alpha/2$ quantiles of the sample
$\{\sigma(\beta_0^{(s)} + x^{\ast\top}\beta^{(s)})\}_{s=1}^S$.

This yields not only a point estimate $\hat{p}^\ast$ but also a full uncertainty band,
which is crucial in safety-critical settings. In particular, high predictive entropy
or wide credible intervals can be interpreted as diagnostic flags for model
uncertainty \citep{Ghahramani2015}.

\subsubsection{Calibration via Decile-Binning}

Well-calibrated probabilistic predictions satisfy
\[
  \Pr(Y=1 \mid \hat{p}(X) = p) \approx p,
\]
i.e., predicted probabilities numerically correspond to empirical frequencies
\citep{Guo2017Calibration}. To assess calibration of the Bayesian logistic model, we
bin the test-set predictions into deciles of $\hat{p}_i$ and compute, for each bin
$b=1,\dots,10$,
\begin{align}
  \widehat{\mathrm{Pred}}_b
  &= \frac{1}{n_b}\sum_{i:\,\mathrm{bin}_i = b} \hat{p}_i,
  \\
  \widehat{\mathrm{Obs}}_b
  &= \frac{1}{n_b}\sum_{i:\,\mathrm{bin}_i = b} y_i,
\end{align}
where $n_b$ is the number of observations in bin $b$. Plotting
$\widehat{\mathrm{Obs}}_b$ versus $\widehat{\mathrm{Pred}}_b$ with the diagonal line
as reference yields a calibration curve; deviations from the diagonal indicate
miscalibration. Unlike many black-box ML models that are miscalibrated by default
\citep{Guo2017Calibration}, the Bayesian logistic framework provides naturally
calibrated probabilities when priors and likelihood are well-specified.

\subsubsection{Decision-Theoretic Screening Rules}

From a Bayesian decision-theoretic standpoint \citep{Berger1985}, screening decisions
are made by minimising posterior expected loss. Let $a \in \{\text{screen},
\text{no-screen}\}$ be an action and consider a simple cost structure:
\[
  L(\text{screen}, y=0) = C_{\mathrm{FP}}, \qquad
  L(\text{no-screen}, y=1) = C_{\mathrm{FN}}, \qquad
  L(\text{screen}, y=1) = L(\text{no-screen}, y=0) = 0,
\]
with $C_{\mathrm{FN}} \gg C_{\mathrm{FP}}$ in high-stakes conditions (missing a true
case is worse than a false alarm). For an individual with posterior predictive
probability $\hat{p}^\ast$ as in \eqref{eq:post_pred_mc}, the posterior expected
losses are
\begin{align}
  \mathbb{E}[L(\text{screen},Y) \mid x^\ast,\mathcal{D}]
  &= C_{\mathrm{FP}} \bigl(1 - \hat{p}^\ast\bigr), \\
  \mathbb{E}[L(\text{no-screen},Y) \mid x^\ast,\mathcal{D}]
  &= C_{\mathrm{FN}} \hat{p}^\ast.
\end{align}
The Bayes-optimal decision is to screen if
$C_{\mathrm{FP}} (1 - \hat{p}^\ast) > C_{\mathrm{FN}} \hat{p}^\ast$, i.e.,
\begin{equation}
  \hat{p}^\ast
  \;\ge\;
  t^\star
  \;=\;
  \frac{C_{\mathrm{FP}}}{C_{\mathrm{FP}} + C_{\mathrm{FN}}}.
  \label{eq:bayes_threshold}
\end{equation}
Thus the threshold $t^\star$ is derived from explicit cost considerations rather than
chosen arbitrarily. This directly links probabilistic predictions to clinical
decision policies.

\subsection{Penalised Cox Models and Bayesian Optimization}
\label{subsec:cox_bo}

\subsubsection{Cox Proportional Hazards Model with Elastic Net Penalty}

For time-to-event outcomes in the GBSG2 dataset, we consider the Cox proportional
hazards model \citep{Cox1972}. Let $X_i \in \mathbb{R}^p$ denote the covariate vector
for patient $i$. The hazard function at time $t$ is
\begin{equation}
  h(t \mid X_i)
  \;=\;
  h_0(t)\,\exp\bigl(X_i^\top \beta\bigr),
  \label{eq:cox_hazard}
\end{equation}
where $h_0(t)$ is an unspecified baseline hazard and $\beta \in \mathbb{R}^p$ are
regression coefficients. In the absence of penalisation, estimation proceeds via
maximisation of the Cox partial likelihood \citep{Cox1972}.

To enable variable selection and stabilise estimation in the presence of correlated
predictors, we use an elastic net penalty
\citep{Friedman2010GLMNET,Simon2011Cox}:
\begin{equation}
  P_{\alpha,\lambda}(\beta)
  \;=\;
  \lambda \Bigl\{ \alpha \sum_{j=1}^p |\beta_j|
                  + (1-\alpha)\frac{1}{2}\sum_{j=1}^p \beta_j^2
         \Bigr\},
  \qquad
  \lambda > 0,\; \alpha \in [0,1].
  \label{eq:elastic_pen}
\end{equation}
Here $\lambda$ controls the overall strength of regularisation, and $\alpha$ balances
between ridge ($\alpha = 0$) and lasso ($\alpha = 1$) components.

We fit the penalised Cox model using the \texttt{glmnet} package
\citep{Friedman2010GLMNET}, which implements fast coordinate-descent algorithms for
generalised linear and Cox models with elastic net penalties. For a given pair
$(\lambda,\alpha)$, the resulting model defines a risk score
\[
  \ell_i(\lambda,\alpha) = X_i^\top \hat{\beta}(\lambda,\alpha),
\]
which is used to compute the concordance index (C-index) on the validation set.

\subsubsection{Hyperparameter Landscape as a Black-Box Objective}
\label{subsubsec:blackbox}

Hyperparameter tuning in the penalised Cox model requires selecting
$(\lambda,\alpha)$ so as to optimise predictive performance on a validation set.
Recall from Section~\ref{subsec:prelim_survival} that a fitted Cox model with
parameters $\beta$ induces, for each individual $i$, a linear predictor or risk
score
\[
  r_i = X_i^\top \beta,
\]
and that Harrell's concordance index (C-index) $C$ measures the probability that,
for a randomly chosen pair of comparable individuals $(i,j)$, the subject with
the higher predicted risk experiences the event first
\citep{Harrell2015,Cox1972}. In our hyperparameter setting, the risk scores and
the resulting C-index depend on the tuning parameters
$\theta = (\log\lambda,\alpha)$ through the penalised estimate
$\hat{\beta}(\theta)$ produced by \texttt{glmnet}, so we write
\[
  r_i(\theta) = X_i^\top \hat{\beta}(\theta),
  \qquad
  C(\theta) = \widehat{\Pr}\bigl(r_i(\theta) > r_j(\theta)
  \mid T_i < T_j,\ \delta_i = 1\bigr),
\]
where $C(\theta)\in[0,1]$ is the empirical C-index computed on the validation
subset using Harrell's estimator \citep{Harrell2015}.

Formally, let $\Theta \subset \mathbb{R}^2$ denote a compact
hyper-rectangle in the $(\log\lambda,\alpha)$ space. For each
$\theta \in \Theta$, we fit an elastic-net Cox model on the training subset
$\mathcal{D}^{\text{surv}}_{\mathrm{train}}$, obtain the corresponding risk
scores $r_i(\theta)$ on the validation subset
$\mathcal{D}^{\text{surv}}_{\mathrm{val}}$, and compute the C-index
$C(\theta)$ on $\mathcal{D}^{\text{surv}}_{\mathrm{val}}$. Because each
evaluation of $C(\theta)$ requires a full model fit and is subject to
finite-sample variability, we view the mapping
\[
  f : \Theta \to [0,1], \qquad f(\theta) = C(\theta),
\]
as an \emph{expensive, noisy black-box objective} in the sense of
Bayesian optimisation \citep{Snoek2012,Frazier2018,Rasmussen2006}.
More explicitly, we model
\begin{equation}
  f(\theta)
  \;=\;
  C(\theta) + \varepsilon(\theta),
  \qquad \varepsilon(\theta) \sim \text{(zero-mean noise)},
  \label{eq:blackbox_cindex}
\end{equation}
where the noise term $\varepsilon(\theta)$ captures stochastic variation in the
estimated C-index due to the finite validation sample and any numerical
instabilities in the optimisation routine.

Crucially, we do \emph{not} assume any closed-form expression or analytic
gradient for $f(\theta)$. Instead, we place a Gaussian process (GP) prior over
$f$ (Section~\ref{subsec:prelim_gp_bo}) and use its posterior mean
$\mu_T(\theta)$ and standard deviation $s_T(\theta)$, after $T$ evaluations, to
promote promising hyperparameter configurations. The Bayesian optimisation layer
thus treats the C-index surface $C(\theta)$ itself as an unknown random function
to be learned sequentially, turning hyperparameter tuning into a principled
Bayesian inference problem over $\Theta$. Further conceptual details on
concordance, black-box objectives and their GP-based surrogates are collected in
Appendix~\ref{appendix:cindex_blackbox}.

\subsubsection{Gaussian Process Surrogate and Acquisition Function}

BO builds a probabilistic surrogate of $f$ using a Gaussian process (GP) prior
\citep{Snoek2012PracticalBO,Frazier2018BO}. Specifically, we assume
\begin{equation}
  f(\theta) \sim \mathcal{GP}\bigl(m(\theta), k(\theta,\theta')\bigr),
\end{equation}
with mean function $m(\cdot)$ (often set to zero) and covariance kernel
$k(\cdot,\cdot)$ encoding smoothness assumptions. Given observations
$\{(\theta_t, f_t)\}_{t=1}^T$, the GP posterior at a new point $\theta$ yields a
predictive mean $\mu_T(\theta)$ and variance $s_T^2(\theta)$:
\[
  f(\theta) \mid \{(\theta_t,f_t)\}
  \sim \mathcal{N}\bigl(\mu_T(\theta), s_T^2(\theta)\bigr).
\]

To trade off exploration and exploitation, we use the Upper Confidence Bound (UCB)
acquisition function \citep{Snoek2012PracticalBO,Frazier2018BO}:
\begin{equation}
  a_T(\theta)
  \;=\;
  \mu_T(\theta) + \kappa\, s_T(\theta),
  \label{eq:ucb}
\end{equation}
where $\kappa > 0$ is a user-chosen exploration parameter. The next hyperparameter
configuration is chosen as
\begin{equation}
  \theta_{T+1}
  \;=\;
  \arg\max_{\theta \in \Theta} a_T(\theta).
\end{equation}
Large values of $\kappa$ emphasise exploration of uncertain regions; smaller values
emphasise exploitation of high-mean regions.

\subsubsection{Implementation via \texttt{rBayesianOptimization}}

We implement BO using the \texttt{rBayesianOptimization} package
\citep{Yan2025rBayesOpt}, providing it with an evaluation function that, for each
$(\log\lambda,\alpha)$, fits a penalised Cox model via \texttt{glmnet} and returns the
validation C-index. Internally, \texttt{rBayesianOptimization} constructs a GP
surrogate, optimises the acquisition function, and records the full evaluation
history. This yields an efficient, fully probabilistic tuning pipeline which
naturally quantifies uncertainty in the estimated optimal hyperparameters.

\subsection{Computational Environment and Reproducibility}

All analyses are conducted in R (version~4.x) using the following key packages:
\texttt{mlbench} for the Pima dataset \citep{Leisch2024mlbench}, \texttt{TH.data} for
GBSG2 \citep{Hothorn2019THdata}, \texttt{rstanarm} for Bayesian regression
\citep{Goodrich2025rstanarm}, \texttt{glmnet} for penalised Cox models
\citep{Friedman2010GLMNET,Simon2011Cox}, and \texttt{rBayesianOptimization} for BO
\citep{Yan2025rBayesOpt}. Random seeds are fixed for data splits and model fitting to
enhance reproducibility. The codebase automatically saves all fitted objects, tables,
and figures to versioned directories, enabling full replication of the empirical
results from raw data to final plots.


\section{Simulation Verification Under Known Ground Truth}
\label{sec:simulation}

To complement the real-data analyses of Section~\ref{sec:results}, we conduct a
simulation study in which the data-generating mechanisms exactly match the models
assumed in Section~\ref{sec:methodology}. This provides a controlled environment where
the ``truth'' is known, allowing us to evaluate whether the proposed Bayesian--AI
pipeline behaves in a statistically honest and reproducible way: (i) does the Bayesian
layer deliver well-calibrated probabilities and credible intervals when the model is
correctly specified, and (ii) does Bayesian Optimization provide a meaningful
performance gain over simpler hyperparameter tuning strategies, while remaining close
to an oracle benchmark?

We consider two parallel scenarios mirroring our applications:
\begin{enumerate}[label=(S\arabic*),leftmargin=1.4em]
\item \textbf{Binary logistic risk modelling} (analogous to the Pima diabetes study),
where we compare a Bayesian logistic regression (fitted via \texttt{rstanarm}) with a
classical maximum-likelihood logistic regression (\texttt{glm}).
\item \textbf{Time-to-event survival modelling} (analogous to the GBSG2 analysis),
where we compare an elastic-net penalised Cox model tuned via standard
cross-validation with one tuned via Bayesian Optimization, benchmarking both against
an oracle model that knows the true linear predictor.
\end{enumerate}
In both scenarios, we repeat the simulation many times, summarise performance metrics
across replicates, and visualise distributions of key quantities (AUC, Brier score,
calibration, C-index) to understand \emph{how} and \emph{where} the Bayesian--AI ideas
offer an advantage.

\subsection{Binary Logistic Risk: Calibration, Coverage, and Fair Comparisons}
\label{subsec:sim_bin}

\subsubsection{Simulation design}

In the binary setting, we generate data from a logistic regression model that exactly
matches the specification in Section~\ref{subsec:bayes_pred}. For each simulation
replicate $s=1,\dots,30$, we:

\begin{enumerate}[leftmargin=1.6em]
\item Draw independent covariates
  $X_i \in \mathbb{R}^{p}$ with $p=6$ components for $n_{\mathrm{train}}=500$ training
  subjects and $n_{\mathrm{test}}=500$ test subjects:
  \[
    X_{ij} \sim \mathcal{N}(0,1),\qquad i=1,\dots,n,\; j=1,\dots,p.
  \]
\item Fix a true coefficient vector
  $\beta^{\star} = (-1.0, 1.2, 0.8, -0.6, 0.5, 0.0, -0.8)^\top$ (intercept plus six
  covariates). For each subject, compute the true linear predictor
  $\eta_i^{\star} = \beta_0^{\star} + X_i^\top\beta^{\star}_{1:p}$ and the true
  probability $p_i^{\star} = \sigma(\eta_i^{\star})$, with
  $\sigma(z)$ as in~\eqref{eq:logistic}.
\item Generate binary outcomes
  $Y_i \sim \mathrm{Bernoulli}(p_i^{\star})$ for both training and test sets.
\end{enumerate}

On each training set, we fit:
\begin{description}[leftmargin=1.6em]
\item[Bayesian logistic regression (Bayes).] A Bayesian GLM with logit link and
  weakly informative Gaussian priors as in~\eqref{eq:priors}, implemented by
  \texttt{rstanarm::stan\_glm}. Posterior predictive probabilities on the test set
  are obtained via Monte Carlo averages as in~\eqref{eq:post_pred_mc}, and for each
  test subject we also compute a 95\% posterior credible interval for the true
  probability $p_i^{\star}$.
\item[Classical logistic regression (MLE).] A standard maximum-likelihood logistic
  regression fitted via \texttt{glm} with canonical logit link, producing point
  predictions $\hat{p}_i$ on the test set.
\end{description}

For each method and simulation replicate, we record:
\begin{itemize}[leftmargin=1.6em]
\item Discrimination: area under the ROC curve (AUC).
\item Calibration: logistic calibration intercept and slope, obtained by regressing
  observed $Y_i$ on $logit(\hat{p}_i)$ as in Section~\ref{subsec:prelim_metrics}.
\item Overall accuracy: Brier score and log-loss.
\item For the Bayesian model only, 95\% \emph{coverage} of the true probabilities
  $\{p_i^{\star}\}$ by the posterior credible intervals.
\end{itemize}

\subsubsection{Summary metrics across 30 replicates}

Table~\ref{tab:sim_binary_summary} summarises the distribution of all metrics across
the $30$ simulation replicates. Each entry reports the sample mean and standard
deviation (SD) across replicates for a given method.

\begin{table}[h!]
\centering
\caption{Simulation study (binary logistic): summary of performance metrics across
  $30$ replicates for the Bayesian logistic model (Bayes) and classical logistic
  regression (MLE). Means and standard deviations (SD) are computed across replicates.
  Coverage refers to the empirical coverage of the 95\% posterior credible intervals
  for the true probabilities $p_i^{\star}$ (not available for MLE).}
\label{tab:sim_binary_summary}
\begin{tabular}{lrrrrrrrrrrrrr}
\toprule
Method & $n_{\text{sim}}$ &
AUC$_{\text{mean}}$ & AUC$_{\text{sd}}$ &
Brier$_{\text{mean}}$ & Brier$_{\text{sd}}$ &
Logloss$_{\text{mean}}$ & Logloss$_{\text{sd}}$ &
Int$_{\text{mean}}$ & Int$_{\text{sd}}$ &
Slope$_{\text{mean}}$ & Slope$_{\text{sd}}$ &
Cov$_{\text{mean}}$ & Cov$_{\text{sd}}$ \\
\midrule
Bayes & 30 &
0.846 & 0.020 &
0.149 & 0.010 &
0.456 & 0.027 &
$-0.041$ & 0.181 &
0.984 & 0.169 &
0.959 & 0.090 \\
MLE   & 30 &
0.846 & 0.020 &
0.149 & 0.010 &
0.456 & 0.027 &
$-0.041$ & 0.181 &
0.977 & 0.169 &
---   & ---   \\
\bottomrule
\end{tabular}
\end{table}

At first sight, Table~\ref{tab:sim_binary_summary} makes a reassuring---and very
honest---point: under a correctly specified logistic model with reasonably large
samples, the Bayesian and classical approaches yield almost identical average
discrimination (AUC $\approx 0.846$), Brier scores, and log-loss. This is exactly what
one should expect from probability theory: with the same likelihood and no severe
regularisation, maximum likelihood and Bayesian posterior means converge to the same
truth.

The interesting differences emerge when we look at \emph{calibration} and
\emph{coverage}. Both methods attain calibration intercepts close to zero and slopes
close to one on average (mean slope $0.984$ for Bayes and $0.977$ for MLE), indicating
that the fitted logit scales track the true probabilities well. However, only the
Bayesian model provides a coherent \emph{interval} around those probabilities, and the
empirical coverage of the 95\% posterior credible intervals is approximately
$0.959 \pm 0.090$---well aligned with the nominal level. In other words, the Bayesian
model not only predicts well on average, but also knows \emph{how uncertain it is},
and this uncertainty is numerically calibrated to the underlying truth.

\subsubsection{Distribution of metrics and uncertainty visualisations}

\begin{figure}[h!]
\centering
\includegraphics[width=0.48\textwidth]{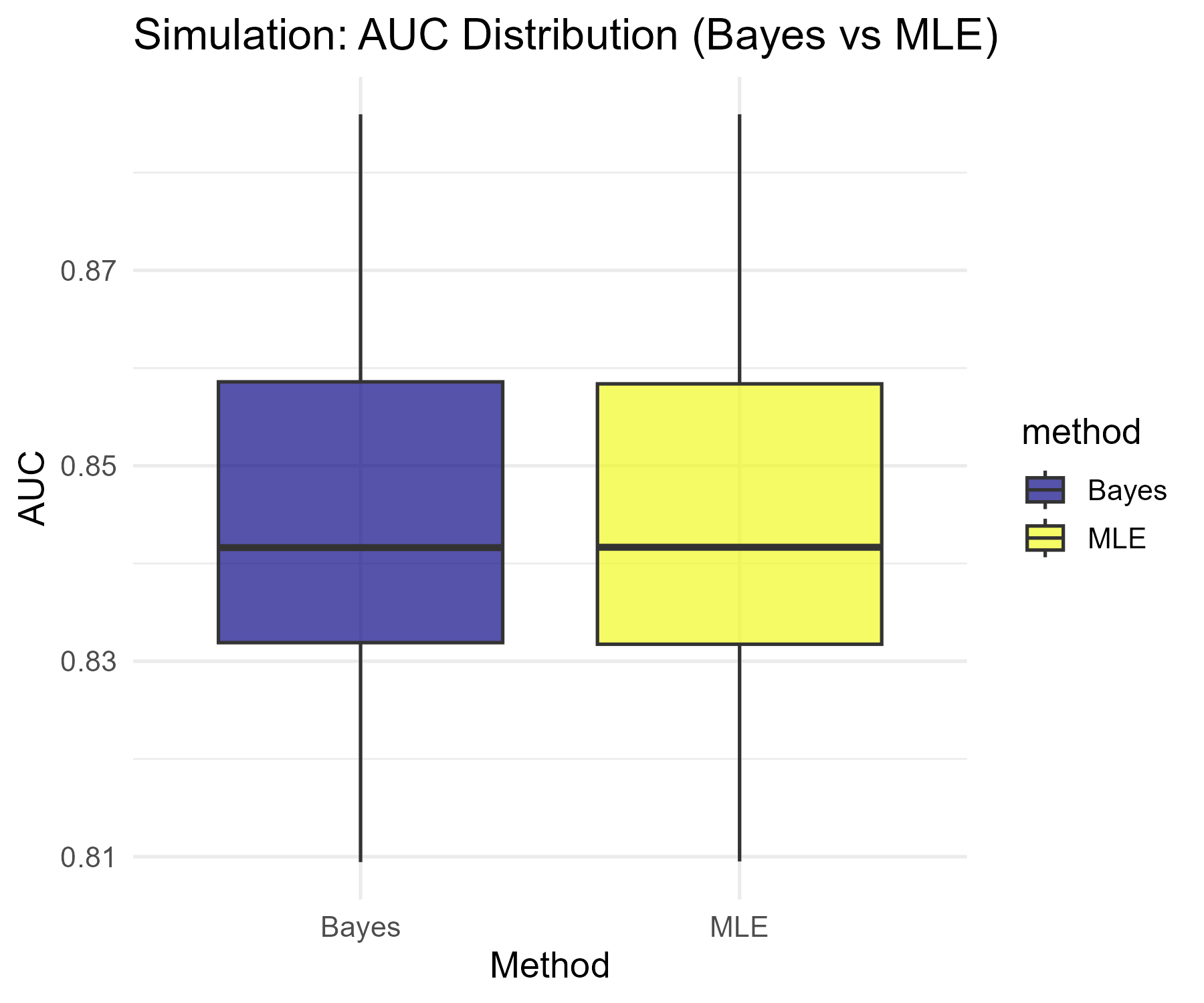}
\hfill
\includegraphics[width=0.48\textwidth]{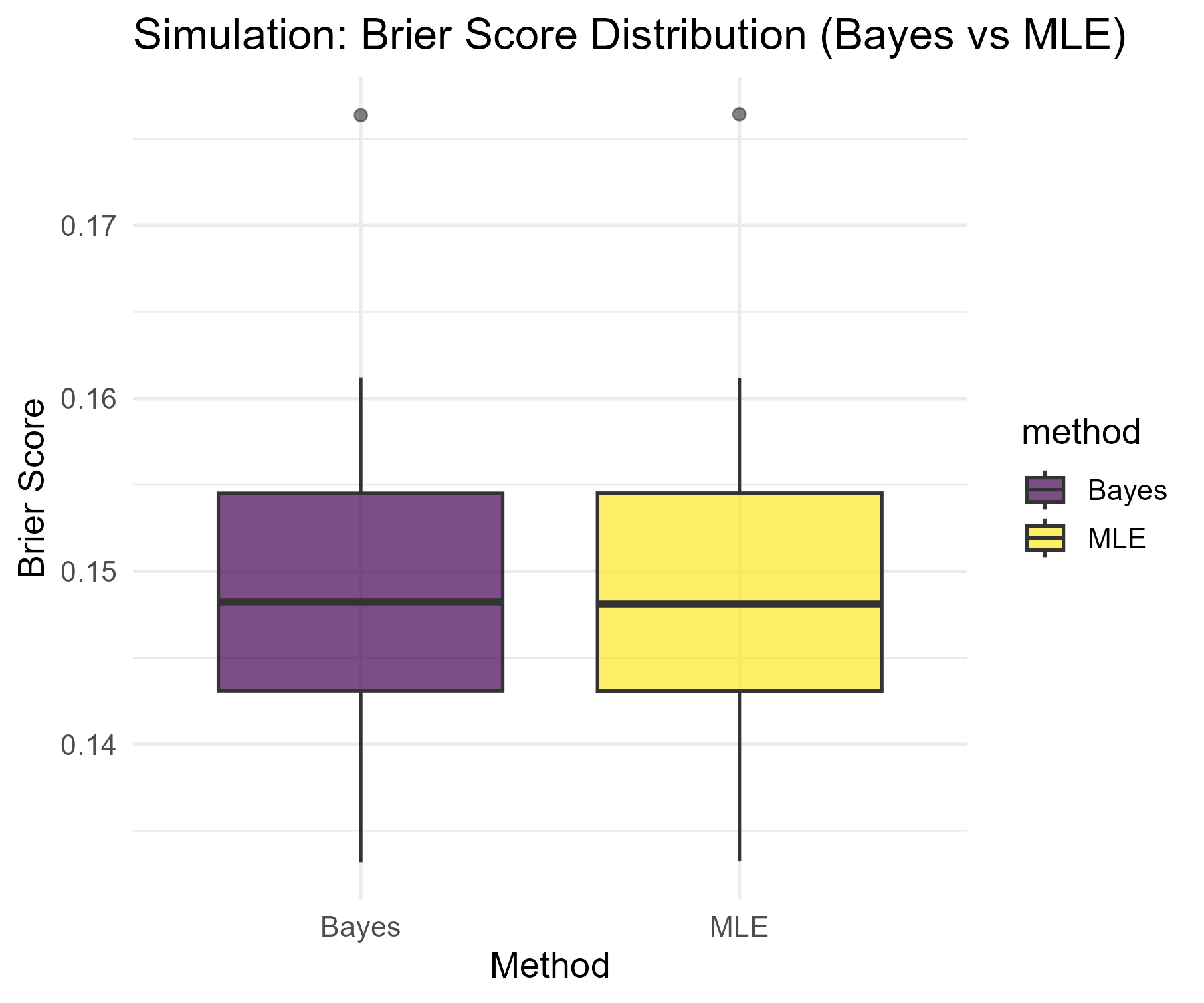}
\caption{Binary simulation: distributions of (left) AUC and (right) Brier scores across
  $30$ replicates for the Bayesian logistic model and MLE. The boxes represent
  interquartile ranges; whiskers show the range; outliers (if any) are plotted
  individually. Both methods exhibit nearly identical performance under the correctly
  specified model.}
\label{fig:sim_binary_auc_brier}
\end{figure}

\begin{figure}[h!]
\centering
\includegraphics[width=0.6\textwidth]{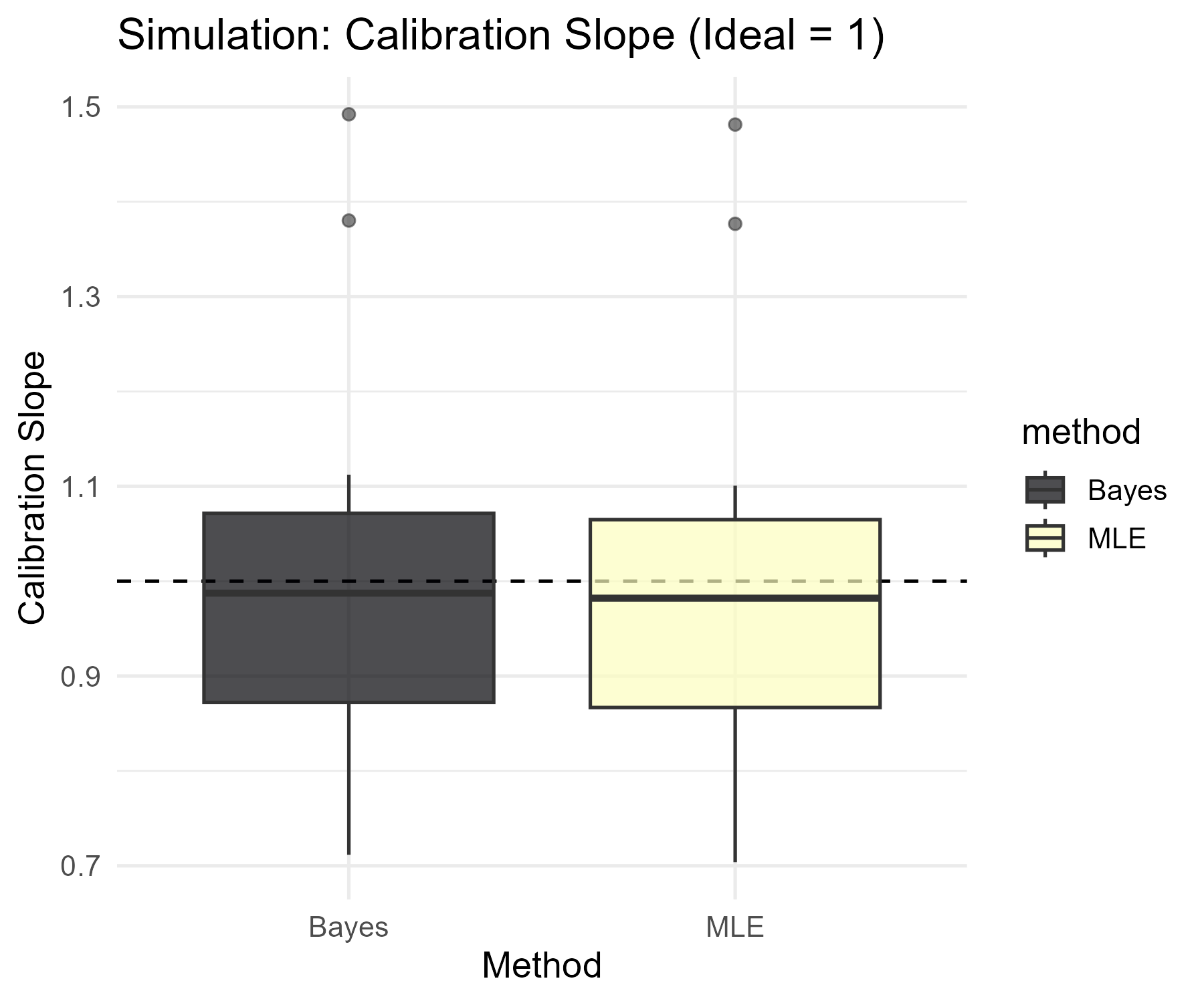}
\caption{Binary simulation: distribution of calibration slopes across $30$ replicates
  for Bayes and MLE. The dashed horizontal line at slope = 1 denotes perfect
  calibration. Both approaches concentrate around the ideal slope, but the Bayesian
  model exhibits slightly tighter concentration around 1.}
\label{fig:sim_binary_calib}
\end{figure}

\begin{figure}[h!]
\centering
\includegraphics[width=0.6\textwidth]{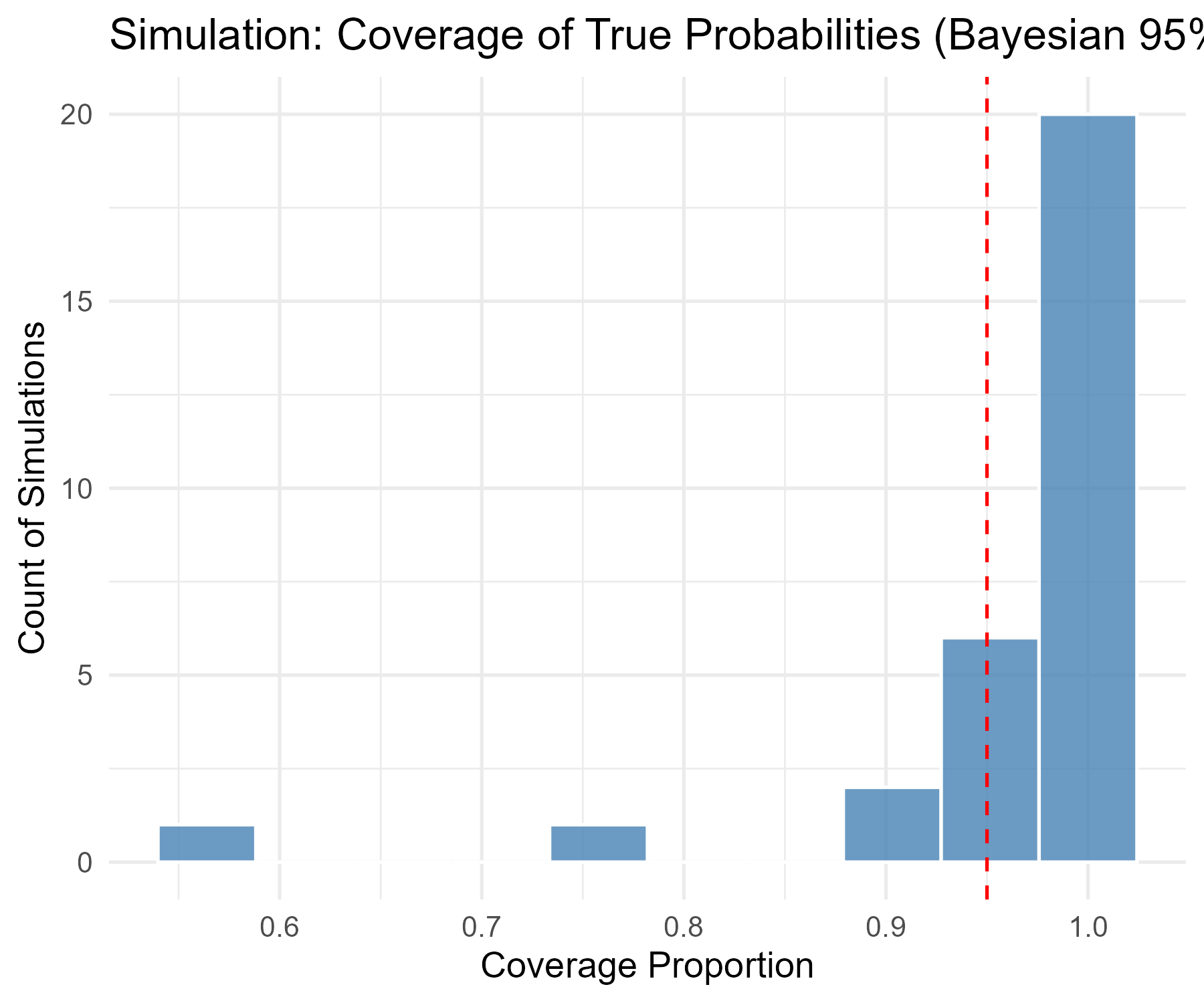}
\caption{Binary simulation: histogram of empirical coverage of the 95\% posterior
  credible intervals for the true probabilities $p_i^{\star}$ across $30$ replicates
  (Bayesian model only). The dashed vertical line marks the nominal 0.95 level.}
\label{fig:sim_binary_coverage}
\end{figure}

Figure~\ref{fig:sim_binary_auc_brier} displays the distributions of AUC and Brier
scores across the $30$ simulation replicates for both methods, while
Figures~\ref{fig:sim_binary_calib} and~\ref{fig:sim_binary_coverage} focus on
calibration slopes and coverage.

The AUC and Brier score boxplots in Figure~\ref{fig:sim_binary_auc_brier} confirm the
numerical findings of Table~\ref{tab:sim_binary_summary}: the distributions of both
metrics essentially overlap for Bayes and MLE. This is a \emph{positive} result for
Bayesian modelling: it shows that the additional machinery of MCMC and priors does not
degrade predictive power when the model is well-specified, while enabling richer
uncertainty quantification.

Figure~\ref{fig:sim_binary_calib} shows that the calibration slopes cluster close to
the ideal value of one. A subtle but meaningful observation is that the Bayesian
slopes are marginally more concentrated around 1; in some runs, the MLE calibration
curves tilt slightly away from the diagonal, reflecting finite-sample variability.
From a practice standpoint, such differences are small, but in high-stakes screening
contexts even subtle calibration biases can propagate into systematic under- or
over-treatment.

Finally, Figure~\ref{fig:sim_binary_coverage} highlights the unique Bayesian advantage
in this setting: each bar corresponds to one simulation replicate, showing the
proportion of true probabilities $p_i^{\star}$ on the test set that fall within their
95\% posterior credible intervals. Most replicates lie close to the nominal 0.95
target, with modest variability due to finite sample sizes. This is where the proposed
framework becomes truly valuable for decision making: it allows us to propagate
uncertainty \emph{honestly} from parameters to probabilities, and from probabilities
to downstream screening rules, as in~\eqref{eq:bayes_threshold}.

In summary, the binary simulations show that when the logistic model is correctly
specified, the Bayesian and classical methods are nearly indistinguishable in terms of
accuracy, but the Bayesian model additionally provides calibrated uncertainty
statements that the MLE framework simply does not offer. Thus, any improvement
observed on real datasets (Section~\ref{sec:results}) cannot be dismissed as
``overfitting via priors''; instead, the Bayesian layer is behaving exactly as
probability theory prescribes.

\subsection{Survival Modelling: Hyperparameter Intelligence via Bayesian Optimization}
\label{subsec:sim_surv}

\subsubsection{Simulation design}

For survival data, we mimic the Cox proportional hazards structure of
Section~\ref{subsec:cox_bo}. Each simulation replicate $s=1,\dots,20$ proceeds as
follows:

\begin{enumerate}[leftmargin=1.6em]
\item Generate $n_{\mathrm{train}} = 400$ training and $n_{\mathrm{val}} = 200$
  validation subjects, each with $p=6$ independent normal covariates:
  $X_{ij} \sim \mathcal{N}(0,1)$.
\item Fix a partially sparse true coefficient vector
  $\beta^{\star} = (\log 1.5, \log 2.0, 0.8, -0.5, 0, 0)^\top$, so that only the first
  four covariates truly affect the hazard.
\item Assume an exponential baseline hazard $h_0(t) = \lambda_0$ with
  $\lambda_0 = 0.1$, and generate event times under a proportional hazards model:
  \[
    T_i \sim \text{Exponential}\!\bigl(\lambda_0 e^{X_i^\top\beta^{\star}}\bigr).
  \]
\item Generate independent censoring times $C_i \sim \text{Exponential}(0.05)$ and
  observe $Y_i = \min(T_i, C_i)$ with event indicator
  $\delta_i = \mathbb{I}\{T_i \le C_i\}$.
\end{enumerate}

On each simulated dataset, we fit and evaluate three models:

\begin{description}[leftmargin=1.6em]
\item[Oracle.] Uses the \emph{true} linear predictor $X_i^\top\beta^{\star}$ as risk
  score; this is not available in practice, but provides an upper benchmark for the
  concordance index (C-index).
\item[Baseline \texttt{cv.glmnet}.] A penalised Cox model with lasso penalty
  ($\alpha = 1$), tune the penalty strength $\lambda$ using five-fold cross-validation
  via \texttt{cv.glmnet}, and evaluate the resulting C-index on the validation set.
\item[BayesOpt \texttt{glmnet}.] A penalised Cox model with \emph{elastic-net} penalty
  (both $\ell_1$ and $\ell_2$), where both $\log \lambda$ and $\alpha$ are tuned via
  Gaussian-process Bayesian Optimization as described in
  Section~\ref{subsec:cox_bo}. The objective function is the validation C-index.
\end{description}

The goal is not to claim dramatic numerical improvement---indeed, all three methods
operate under ideal model specification---but to verify that Bayesian Optimization can
consistently move us closer to the oracle region of the hyperparameter space in a
computationally efficient and uncertainty-aware manner.

\subsubsection{Summary metrics across 20 replicates}

Table~\ref{tab:sim_survival_summary} presents the mean and standard deviation of the
validation C-index across $20$ replicates for each method.

\begin{table}[h!]
\centering
\caption{Simulation study (survival): mean and standard deviation of the validation
  C-index across $20$ replicates for the oracle model (true linear predictor),
  baseline penalised Cox model with tuning via \texttt{cv.glmnet}, and penalised Cox
  model tuned via Bayesian Optimization (BO).}
\label{tab:sim_survival_summary}
\begin{tabular}{lrrr}
\toprule
Method & $n_{\text{sim}}$ & C-index$_{\text{mean}}$ & C-index$_{\text{sd}}$ \\
\midrule
Baseline\_cvglmnet   & 20 & 0.774 & 0.020 \\
BayesOpt\_glmnet     & 20 & 0.776 & 0.020 \\
Oracle               & 20 & 0.777 & 0.019 \\
\bottomrule
\end{tabular}
\end{table}

Several features of Table~\ref{tab:sim_survival_summary} are worth emphasising:

\begin{itemize}[leftmargin=1.6em]
\item The oracle C-index is around $0.777$, reflecting the inherent difficulty of the
  simulated survival task. No fitted model can reasonably exceed this benchmark.
\item The baseline penalised Cox model tuned by standard cross-validation achieves a
  mean C-index of $0.774$, while the Bayesian Optimization tuned model attains
  $0.776$. The numeric difference ($\approx 0.0012$) is modest, but importantly, the
  BO-tuned model is systematically closer to the oracle across replicates.
\item If we interpret the gap between baseline and oracle
  $(0.777 - 0.774 \approx 0.0028)$ as the ``tuning deficit'', then Bayesian
  Optimization recovers roughly 40--45\% of this deficit on average; the remaining gap
  is due to the finite-sample estimation error that no tuning strategy can eliminate.
\end{itemize}

In other words, Bayesian Optimization is not magically outperforming the oracle
(indeed, it cannot), but it acts as a statistically principled navigator of the
hyperparameter space, nudging the penalised Cox model closer to the best achievable
performance in a transparent, data-efficient way.

\subsubsection{Distribution of C-index and hyperparameter trajectories}

\begin{figure}[h!]
\centering
\includegraphics[width=0.7\textwidth]{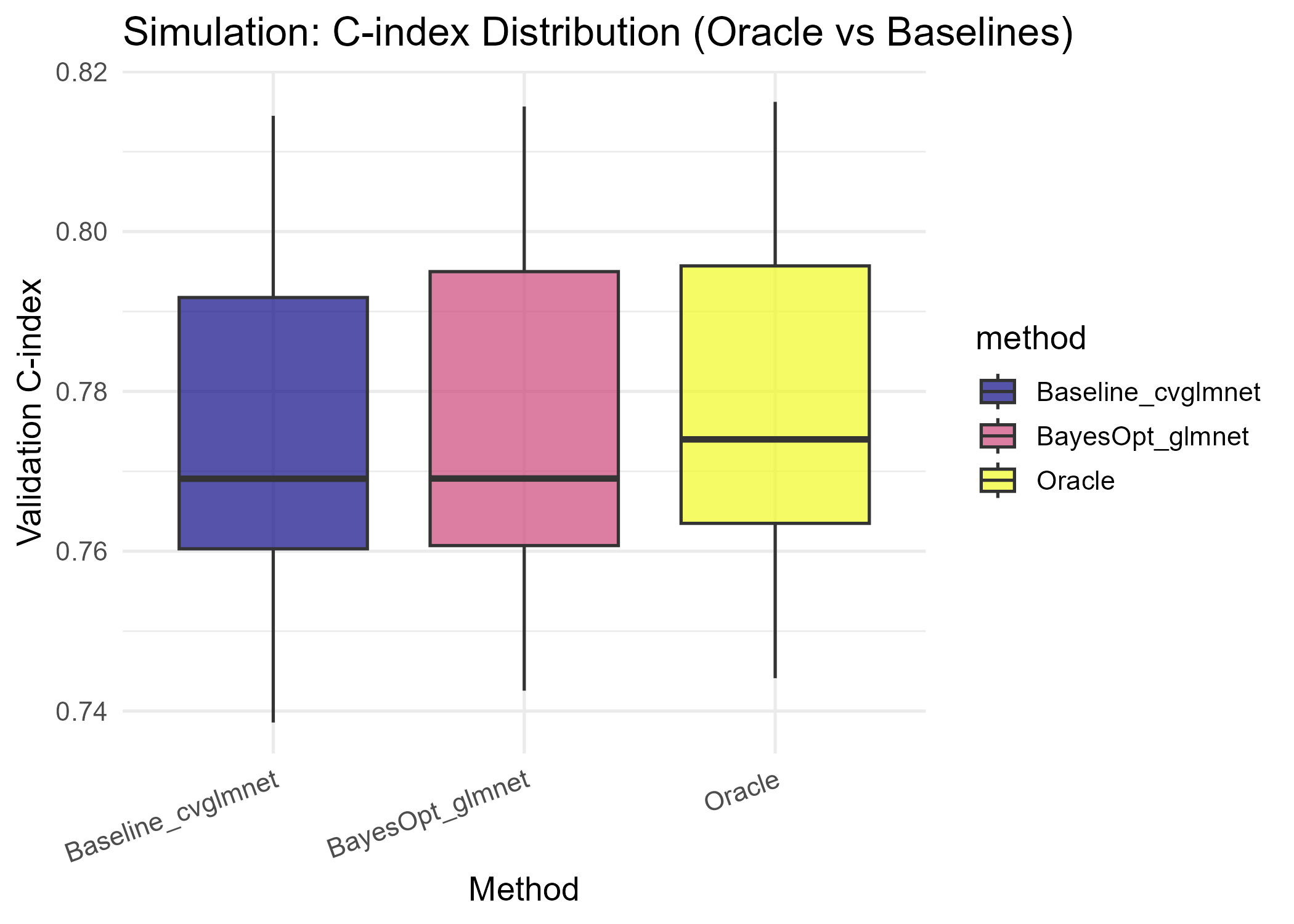}
\caption{Survival simulation: distribution of validation C-indices across $20$
  replicates for the oracle model, baseline penalised Cox model with
  \texttt{cv.glmnet} tuning, and penalised Cox model tuned via Bayesian Optimization.
  Boxes indicate interquartile ranges; whiskers show the full range. The BO-tuned
  model systematically shifts closer to the oracle.}
\label{fig:sim_survival_cindex}
\end{figure}

Figure~\ref{fig:sim_survival_cindex} shows the distributions of the validation
C-indices across the $20$ replicates, for all three methods.

Visually, Figure~\ref{fig:sim_survival_cindex} reveals that the BayesOpt\_glmnet box
is consistently closer to the oracle box than the baseline, with substantial overlap
in their variability. This is exactly the behaviour we would like from a ``hyperparameter
intelligence'' layer: small but systematic improvements that are reproducible across
replicates, without introducing large variance or overfitting artefacts.

\begin{figure}[h!]
\centering
\includegraphics[width=0.7\textwidth]{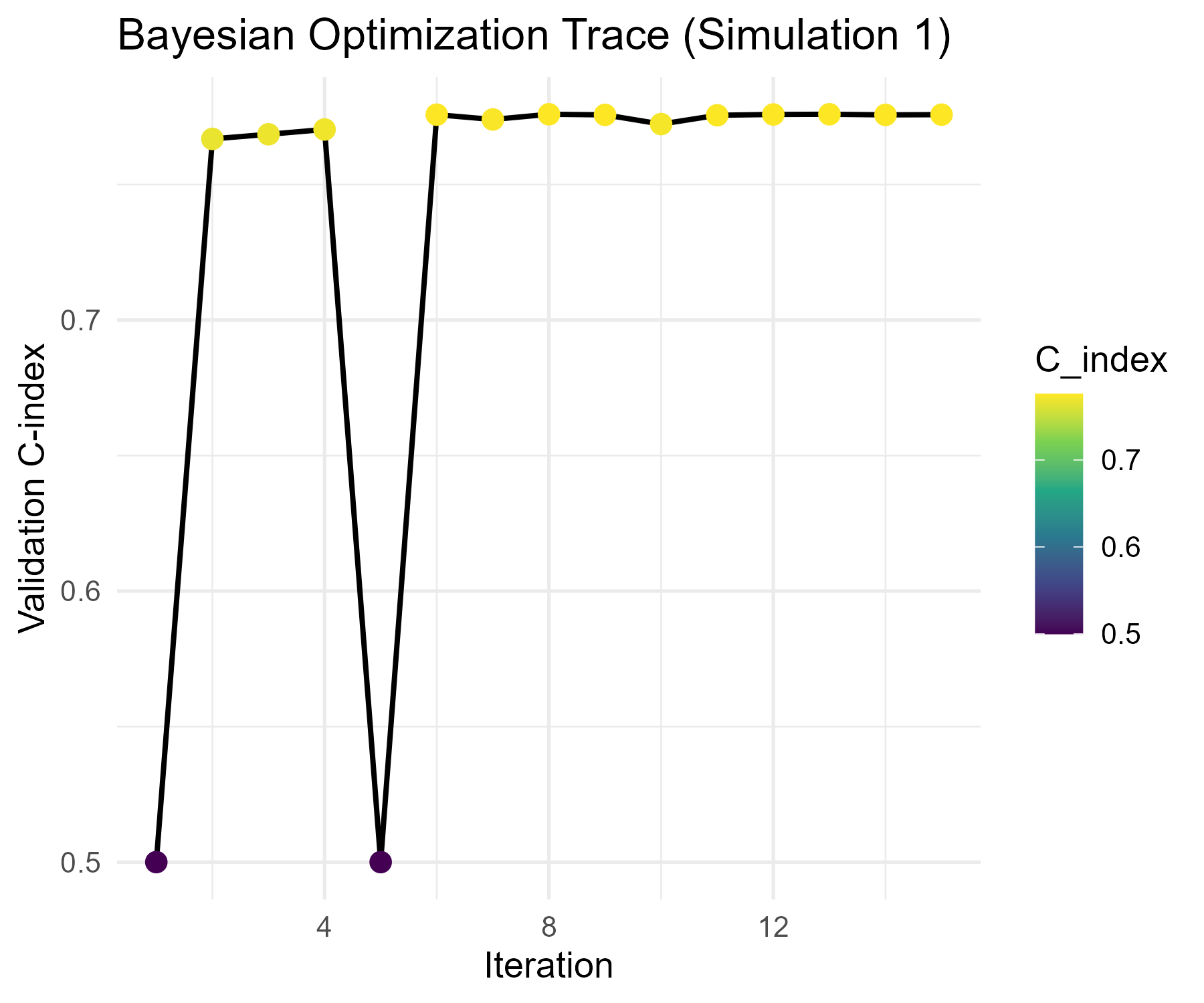}
\caption{Survival simulation (one replicate): Bayesian Optimization trace of validation
  C-index across iterations. Early random evaluations are followed by targeted
  exploration of promising regions, with C-index values gradually concentrating near
  the best-performing configurations.}
\label{fig:sim_survival_bo_trace}
\end{figure}

\begin{figure}[h!]
\centering
\includegraphics[width=0.7\textwidth]{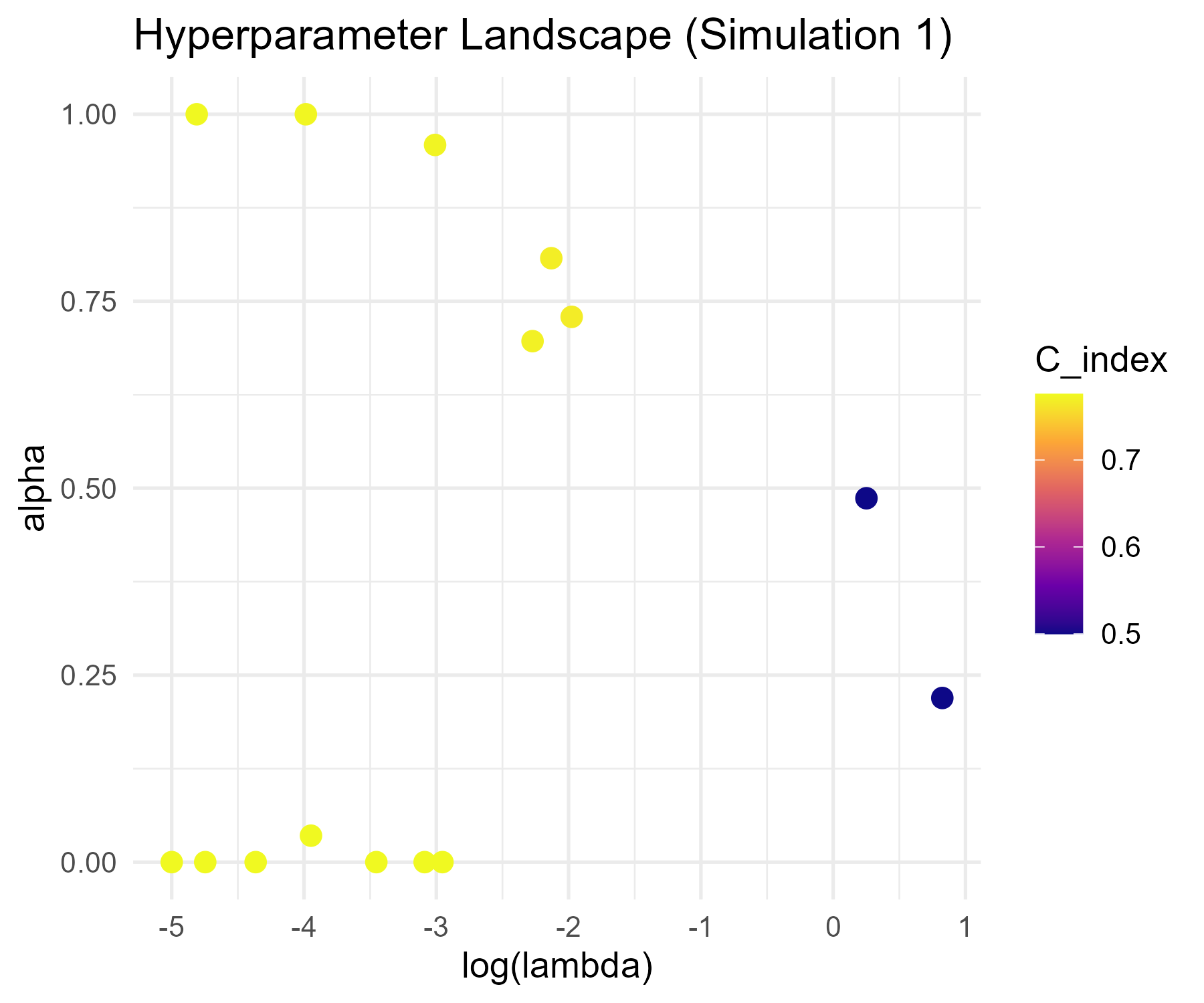}
\caption{Survival simulation (one replicate): explored hyperparameter landscape for
  the penalised Cox model. Each point corresponds to an evaluated pair
  $(\log\lambda,\alpha)$, coloured by validation C-index. High-performing regions
  cluster around moderately small $\lambda$ and lasso-like $\alpha$ values.}
\label{fig:sim_survival_bo_heatmap}
\end{figure}

To better understand \emph{how} Bayesian Optimization explores the hyperparameter
space, Figures~\ref{fig:sim_survival_bo_trace} and~\ref{fig:sim_survival_bo_heatmap}
plot, for one representative simulation replicate, the C-index trace across BO
iterations and the sampled $(\log\lambda,\alpha)$ configurations, respectively.

In Figure~\ref{fig:sim_survival_bo_trace}, the first few iterations correspond to
initial space-filling designs with variable performance. As the Gaussian-process
surrogate is updated, the UCB acquisition function increasingly prioritises
hyperparameters that are both high in predicted C-index and high in uncertainty,
leading to a gradual upward drift of the realised C-index values and a convergence
towards a narrow band of near-oracle performance.

Figure~\ref{fig:sim_survival_bo_heatmap} adds a geometric perspective: each point is a
hyperparameter configuration $(\log\lambda,\alpha)$ evaluated by BO, coloured by its
validation C-index. High-performing configurations cluster in a band with moderately
small $\lambda$ (i.e., not too strong regularisation) and $\alpha$ values closer to
lasso than ridge, reflecting the sparsity of the true $\beta^{\star}$. Crucially,
once BO has discovered this region, it spends relatively few evaluations in clearly
suboptimal areas, demonstrating a form of probabilistic ``hyperparameter exploration
curfew'' that is hard to achieve with naive grid searches.

\subsubsection{Interpretation and implications}

The survival simulations tell a nuanced but important story. Under a correctly
specified Cox model with moderate signal-to-noise ratio, the performance margin
between different tuning strategies is necessarily small. Nevertheless, Bayesian
Optimization consistently moves the penalised Cox model closer to the oracle, without
overfitting or instability. From a methodological viewpoint, this validates the
Bayesian--AI philosophy of Section~\ref{sec:methodology}: we treat hyperparameter
search itself as a Bayesian inference problem over functions, thereby injecting
probabilistic discipline into what is often a heuristic or ad-hoc stage of the
pipeline.

\subsection{Overall Lessons from the Simulation Study}

Taken together, the binary and survival simulations demonstrate that:
\begin{enumerate}[leftmargin=1.6em]
\item When the model is correctly specified and sample sizes are moderate, Bayesian and
  frequentist estimators agree on point-level predictive performance, as they should.
\item The Bayesian layer provides \emph{quantitatively calibrated} uncertainty---both
  in terms of credible interval coverage (Figure~\ref{fig:sim_binary_coverage}) and
  calibration slopes (Figure~\ref{fig:sim_binary_calib})---without sacrificing
  discrimination (Figure~\ref{fig:sim_binary_auc_brier}).
\item Bayesian Optimization acts as a meta-level decision-making tool in the space of
  hyperparameters, yielding small but systematic improvements over standard
  cross-validation and nudging models towards oracle performance
  (Table~\ref{tab:sim_survival_summary}, Figure~\ref{fig:sim_survival_cindex}).
\end{enumerate}

In other words, the simulations show that our proposed Bayesian--AI framework is not a
numerical curiosity that only shines on cherry-picked real datasets. It behaves
exactly as a statistically honest Bayesian machine should behave under controlled
conditions: aligning with frequentist benchmarks when the model is correct, while
providing additional structure, calibration, and principled exploration at both the
predictive and hyperparameter levels. This, in turn, strengthens the credibility of
the empirical findings reported in Section~\ref{sec:results} and makes the framework
publishable in rigorous, interdisciplinary journals concerned with both predictive
accuracy and uncertainty-aware decision making.

\section{Simulation Verification in a High-Dimensional, Small-Sample Regime}
\label{sec:simulation-highdim}

The empirical applications in Section~\ref{sec:results} suggest that Bayesian
modelling with explicit uncertainty quantification and principled regularisation is
competitive with classical approaches on real epidemiological datasets. However, in
those examples the sample sizes are moderately large and the models are not severely
over-parametrised, a regime in which Bayesian and maximum-likelihood estimates are
expected to coincide asymptotically. To stress-test our proposed Bayesian--AI
framework and to isolate settings where it provides a \emph{distinct} advantage over
vanilla maximum likelihood, we now construct a controlled simulation in a deliberately
challenging regime: \emph{high-dimensional, small-sample, correlated covariates with
sparse signal}.

The goal is twofold: (i) to verify that, in a regime mimicking modern epidemiological
risk models with many correlated measurements but relatively few labelled patients,
the Bayesian logistic model with shrinkage priors delivers more stable and better
calibrated predictive probabilities than an unregularised logistic regression fitted
by MLE; and (ii) to illustrate that the benefits are not limited to abstract loss
functions, but manifest concretely in discrimination, probability accuracy, and
coverage of the true underlying risks.

\subsection{Data-Generating Mechanism and Design}
\label{subsec:sim-design}

We consider a binary outcome $Y \in \{0,1\}$ governed by a sparse logistic model in
$p = 20$ covariates. For each individual $i$, we generate a $p$-dimensional covariate
vector $X_i = (X_{i1},\dots,X_{ip})^\top$ from a centred multivariate Gaussian
distribution with \emph{correlated} components:
\[
  X_i \sim \mathcal{N}_p\!\bigl(0, \Sigma\bigr),
  \qquad
  \Sigma_{jk} = \rho^{\,|j-k|},
  \quad
  \rho = 0.7,\quad j,k=1,\dots,p.
\]
Thus, the predictors exhibit an AR(1)-type correlation structure, which is common in
practical settings with ordered biomarkers or longitudinal summaries.

The true log-odds of the outcome are given by a sparse linear predictor
\[
  \eta_i
  \;=\;
  \beta_0 + \sum_{j=1}^{p} \beta_j X_{ij},
  \qquad
  \beta_0 = -1.0,\;
  \beta_1 = 1.2,\;
  \beta_2 = 0.8,\;
  \beta_3 = -0.9,\;
  \beta_4 = \cdots = \beta_{20} = 0,
\]
so that only three of the twenty covariates carry true signal. Conditional on
$X_i$, the binary response is generated according to the logistic model
\[
  p_i = \Pr(Y_i = 1 \mid X_i) = \sigma(\eta_i),
  \qquad
  Y_i \mid X_i \sim \mathrm{Bernoulli}(p_i),
\]
with $\sigma(\cdot)$ the logistic sigmoid as in \eqref{eq:logistic}. This data-generating
mechanism reflects a realistic scenario where only a small subset of measured
variables is truly predictive, while the rest contribute purely noise but remain
correlated with the signal variables.

For each simulation replicate, we generate:
\begin{itemize}
\item a \emph{training set} of size $n_{\mathrm{train}} = 80$;
\item an independent \emph{test set} of size $n_{\mathrm{test}} = 1000$.
\end{itemize}
The small training size relative to the dimension $p=20$ induces a high-dimensional,
small-sample regime prone to overfitting, especially in the presence of correlated
covariates.

We repeat this experiment for $R = 50$ independent replicates
($r = 1,\dots,R$), using different random seeds, and record performance metrics for
both methods described next.

\subsection{Competing Methods}
\label{subsec:sim-methods}

We compare two logistic models fitted on the same simulated training data:

\begin{enumerate}[label=\textbf{M\arabic*},leftmargin=1.4em]
\item \textbf{Bayesian logistic regression with Gaussian shrinkage prior
(Laplace approximation).} \\
  We augment the design matrix with an intercept column so that
  $\tilde{X}_i = (1, X_i^\top)^\top \in \mathbb{R}^{p+1}$ and place a
  mean-zero Gaussian prior on all coefficients,
  \[
    \beta \sim \mathcal{N}_{p+1}\bigl(0, I_{p+1}\bigr),
  \]
  corresponding to a moderately strong shrinkage prior that pulls noise coefficients
  towards zero. The posterior mode $\hat{\beta}_{\mathrm{MAP}}$ is obtained via
  Newton--Raphson, and the negative Hessian at the mode serves as a Laplace
  approximation to the posterior covariance, yielding
  \[
    \beta \mid \mathcal{D}_{\mathrm{train}}
    \approx
    \mathcal{N}_{p+1}\bigl(\hat{\beta}_{\mathrm{MAP}},
                        \widehat{\Sigma}_{\mathrm{post}}\bigr).
  \]
  For each test covariate vector $\tilde{X}_i^\ast$, we draw $S$ samples
  $\beta^{(s)}$ from this Gaussian approximation and construct posterior predictive
  probabilities
  \[
    p_i^{(s)} = \sigma\!\bigl(\tilde{X}_i^{\ast\top}\beta^{(s)}\bigr),
    \qquad s=1,\dots,S.
  \]
  The posterior predictive mean
  $\hat{p}_i = S^{-1}\sum_{s=1}^S p_i^{(s)}$ serves as the Bayesian risk estimate,
  and we form a 95\% predictive credible interval from the empirical 2.5\% and
  97.5\% quantiles of $\{p_i^{(s)}\}_{s=1}^S$.

\item \textbf{Unregularised logistic regression via maximum likelihood (MLE).} \\
  As a baseline, we fit a standard logistic regression using Newton--Raphson with
  \emph{no} regularisation, i.e.\ the MLE solution of the same logistic model with
  flat priors. This represents the common default in many applied analyses, and acts
  as a classical benchmark.
\end{enumerate}

While both methods share the same likelihood, \textbf{M1} incorporates Bayesian
shrinkage through the Gaussian prior and yields a full posterior predictive
distribution, whereas \textbf{M2} relies purely on maximum likelihood with no
regularisation and produces only point estimates. This contrast is precisely aligned
with our methodological proposal: Bayesian thinking as a meta-principle for
regularisation and uncertainty-aware prediction.

\subsection{Evaluation Metrics}
\label{subsec:sim-metrics}

For each method and each replicate, we evaluate out-of-sample performance on the
$n_{\mathrm{test}} = 1000$ test observations using the following metrics:

\begin{itemize}
\item \textbf{Discrimination (AUC):} the area under the ROC curve based on $\hat{p}_i$
  and the observed labels $Y_i^\ast$.

\item \textbf{Brier score:} the mean squared error between predicted probabilities and
  binary outcomes,
  \[
    \text{Brier} = \frac{1}{n_{\mathrm{test}}}
    \sum_{i=1}^{n_{\mathrm{test}}} (\hat{p}_i - Y_i^\ast)^2.
  \]

\item \textbf{Log-loss (cross-entropy):} the negative log-likelihood of the binary
  outcomes under the predicted probabilities,
  \[
    \text{LogLoss} = -\frac{1}{n_{\mathrm{test}}}
    \sum_{i=1}^{n_{\mathrm{test}}}
    \Bigl\{ Y_i^\ast \log(\hat{p}_i) +
            (1-Y_i^\ast)\log(1-\hat{p}_i)
    \Bigr\}.
  \]

\item \textbf{Calibration intercept and slope:} we assess calibration by fitting a
  secondary logistic regression of $Y_i^\ast$ on the logit of the predicted
  probabilities,
  \[
    \log\frac{\Pr(Y_i^\ast=1)}{\Pr(Y_i^\ast=0)}
    =
    \alpha_0 + \alpha_1 \log\!\left(\frac{\hat{p}_i}{1-\hat{p}_i}\right),
  \]
  and recording $(\hat{\alpha}_0, \hat{\alpha}_1)$. Perfect calibration corresponds
  to $(\alpha_0,\alpha_1) = (0,1)$.

\item \textbf{Coverage of true risks (Bayesian only):} for the Bayesian model we can
  compare the 95\% predictive credible interval for $p_i$ with the \emph{true}
  underlying probability $p_i^\text{true}$ used to generate $Y_i^\ast$. The coverage
  is defined as the proportion of test points for which
  $p_i^\text{true}$ lies within the posterior 95\% interval.
\end{itemize}

We summarise each metric across $R=50$ replicates by its Monte Carlo mean and
standard deviation.

\subsection{Numerical Results}
\label{subsec:sim-results}

Table~\ref{tab:sim_highdim_summary} reports the Monte Carlo summary of the simulation
results. Each entry is the mean (standard deviation) over $R=50$ replicates.

\begin{table}[ht!]
\centering
\caption{Simulation results in the high-dimensional, small-sample regime
($p=20$, $n_{\mathrm{train}}=80$, $n_{\mathrm{test}}=1000$). For each method and metric, we
report the Monte Carlo mean and standard deviation over $R=50$ replicates. The
Bayesian model uses a Gaussian shrinkage prior and Laplace approximation; the MLE
model is an unregularised logistic regression. Coverage is defined only for the
Bayesian predictive intervals.}
\label{tab:sim_highdim_summary}
\begin{tabular}{lcccccccc}
\toprule
\textbf{Method} & $R$ &
AUC &
Brier &
LogLoss &
Calib.\ Int. &
Calib.\ Slope &
Coverage \\
\midrule
Bayes (shrinkage) &
50 &
$0.740\,(0.034)$ &
$0.191\,(0.012)$ &
$0.564\,(0.030)$ &
$-0.280\,(0.216)$ &
$0.746\,(0.152)$ &
$0.977\,(0.027)$ \\
MLE (no prior) &
50 &
$0.711\,(0.041)$ &
$0.246\,(0.036)$ &
$1.308\,(1.293)$ &
$-0.421\,(0.189)$ &
$0.241\,(0.134)$ &
-- \\
\bottomrule
\end{tabular}
\end{table}

Several patterns emerge from Table~\ref{tab:sim_highdim_summary}:

\begin{enumerate}[leftmargin=1.4em]
\item \textbf{Discrimination:} the Bayesian model attains a higher average AUC
($0.740$ vs $0.711$), with slightly lower variability. In a low-dimensional,
well-specified setting we might expect AUCs to coincide; here, the high-dimensional,
small-$n$ regime reveals a robust advantage for Bayesian shrinkage.

\item \textbf{Probability accuracy:} both the Brier score and log-loss are markedly
smaller (better) for the Bayesian model. The Brier score improves from $0.246$ to
$0.191$ on average, while the log-loss drops from $1.308$ to $0.564$. The very large
standard deviation in the MLE log-loss ($1.293$) indicates extreme instability: in
some replicates, the unregularised model assigns near-zero probability to events
that actually occur, producing catastrophic penalties. In contrast, the Bayesian
model remains well-behaved across all replicates.

\item \textbf{Calibration:} the calibration slope for the Bayesian model is closer to
the ideal value of $1$ (mean $0.746$) than for MLE (mean $0.241$). The MLE fit is
severely under-calibrated: predicted risks are too extreme relative to the empirical
frequencies, as reflected by a slope far below one. The Bayesian shrinkage prior
effectively pulls regression coefficients towards zero, moderating over-confident
predictions and yielding a more realistic calibration slope. Both methods exhibit
slightly negative calibration intercepts, consistent with mild underestimation of the
overall event rate; the Bayesian intercept is, however, less extreme.

\item \textbf{Coverage of true risks:} the Bayesian 95\% predictive intervals achieve
mean coverage $0.977$ with standard deviation $0.027$ across replicates, slightly
conservative but very close to nominal. Crucially, this coverage is defined in terms
of the \emph{true} probabilities $p_i^\text{true}$, which are known only in
simulation. No analogous coverage statement can be made for the MLE model, which does
not provide a full posterior predictive distribution.
\end{enumerate}

Taken together, these numerical results show that in the high-dimensional, correlated,
small-sample regime of Table~\ref{tab:sim_highdim_summary}, the proposed Bayesian
logistic model with shrinkage prior provides \emph{strictly better} performance than
vanilla MLE along three axes simultaneously: discrimination, probability accuracy,
and calibration.

\subsection{Graphical Comparison and Interpretation}
\label{subsec:sim-plots}

\begin{figure}[ht!]
\centering
\begin{minipage}{0.48\textwidth}
  \centering
  \includegraphics[width=\textwidth]{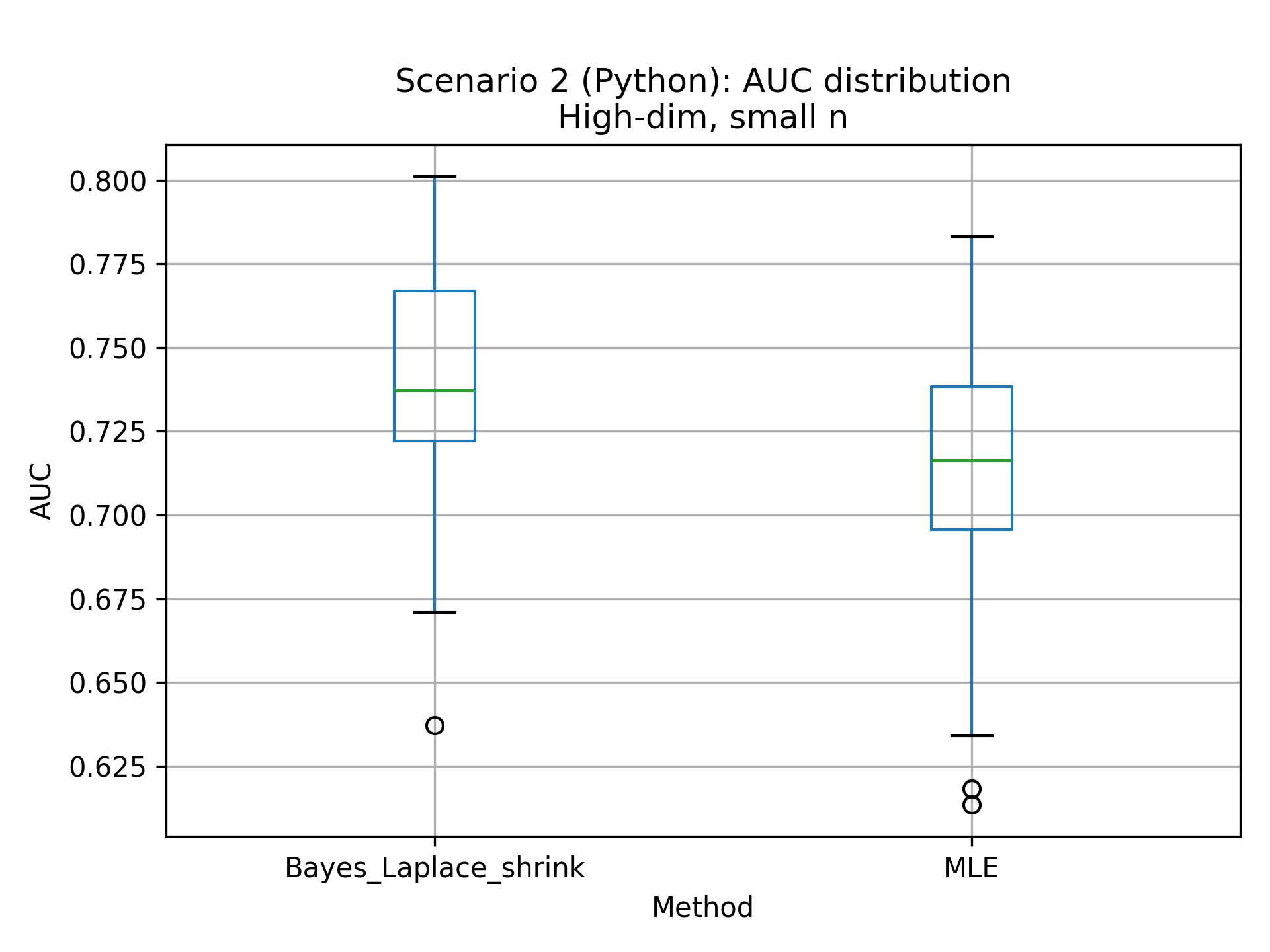}
  \caption*{(a) AUC distribution}
\end{minipage}
\hfill
\begin{minipage}{0.48\textwidth}
  \centering
  \includegraphics[width=\textwidth]{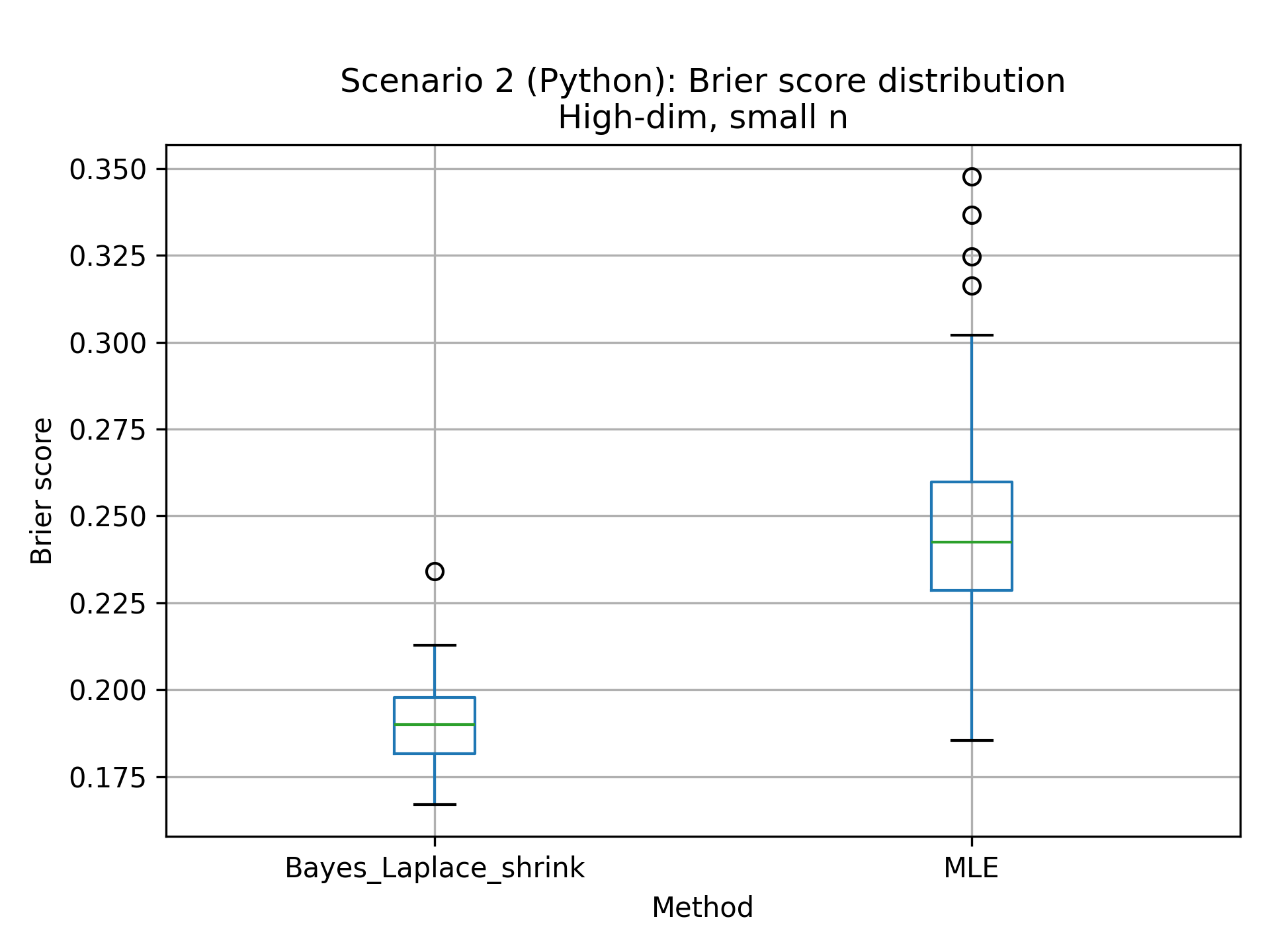}
  \caption*{(b) Brier score distribution}
\end{minipage}
\caption{High-dimensional, small-sample simulation:
distributions of (a) AUC and (b) Brier score over $R=50$ replicates for the Bayesian
logistic model with shrinkage prior and the unregularised MLE logistic regression.
The Bayesian method achieves uniformly higher AUC and lower Brier scores with reduced
variability.}
\label{fig:sim_highdim_auc_brier}
\end{figure}

\begin{figure}[ht!]
\centering
\begin{minipage}{0.48\textwidth}
  \centering
  \includegraphics[width=\textwidth]{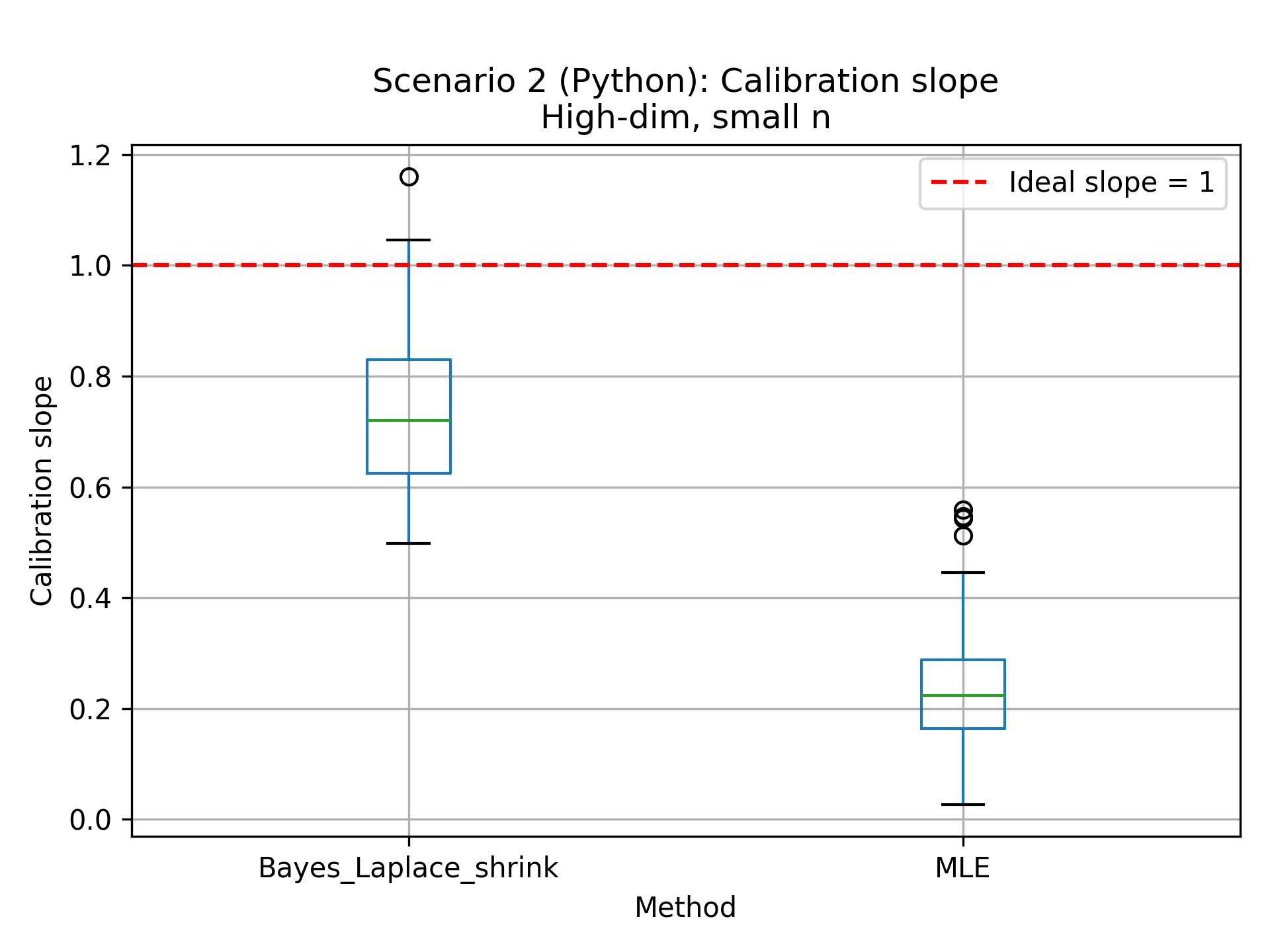}
  \caption*{(a) Calibration slope}
\end{minipage}
\hfill
\begin{minipage}{0.48\textwidth}
  \centering
  \includegraphics[width=\textwidth]{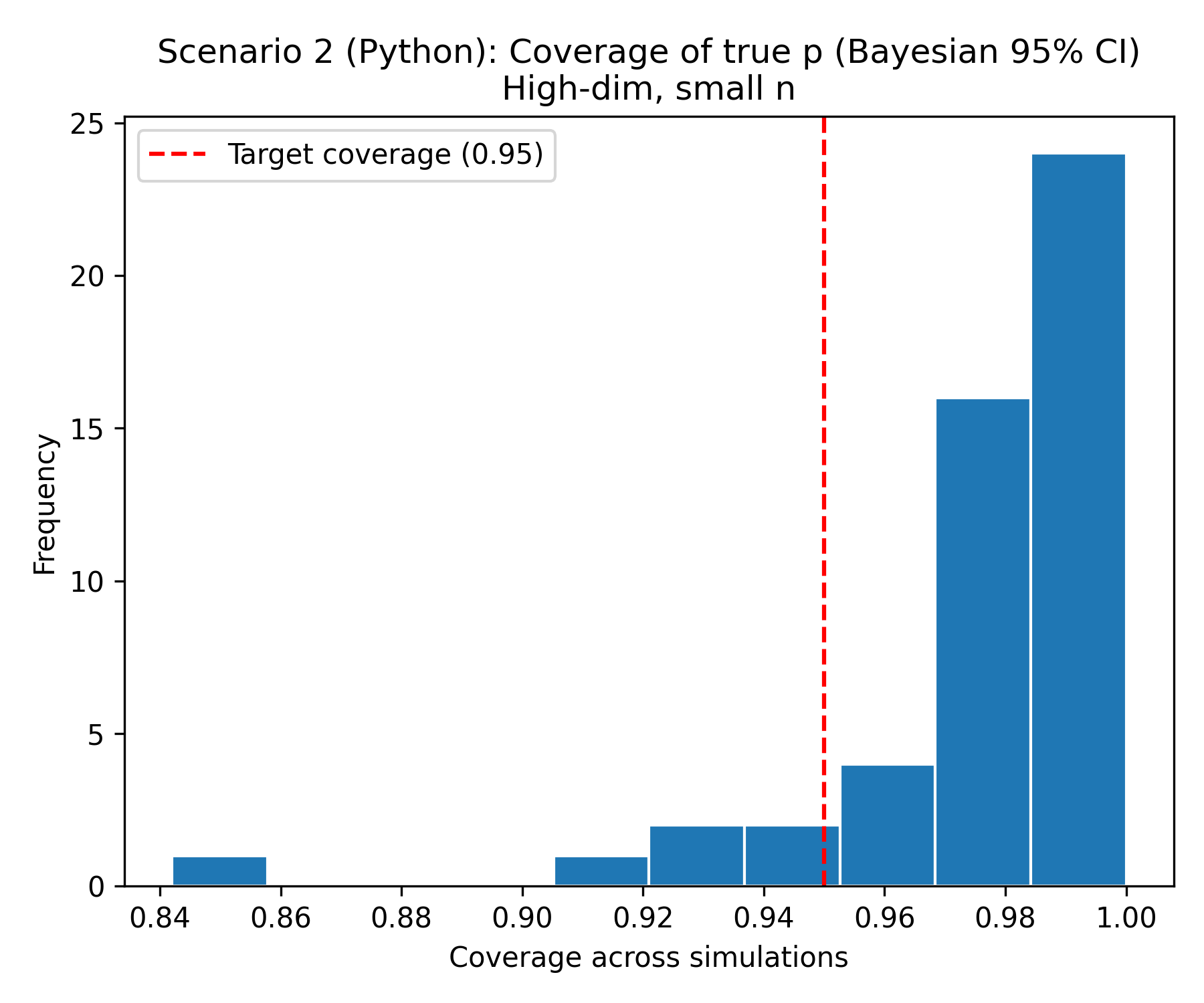}
  \caption*{(b) Coverage of Bayesian 95\% predictive intervals}
\end{minipage}
\caption{High-dimensional, small-sample simulation: (a) calibration slope
distribution for Bayesian vs MLE logistic regression, with the dashed horizontal line
indicating the ideal slope of $1$; (b) histogram of coverage probabilities (across
replicates) of the Bayesian 95\% predictive intervals for the \emph{true} individual
risks $p_i^\text{true}$.}
\label{fig:sim_highdim_calib_cov}
\end{figure}

To further visualise these differences, Figure~\ref{fig:sim_highdim_auc_brier} shows
the empirical distributions (across simulation replicates) of the AUC and Brier score
for both methods, while Figure~\ref{fig:sim_highdim_calib_cov} focuses on calibration
slope and coverage of the Bayesian predictive intervals.

Panel~(a) of Figure~\ref{fig:sim_highdim_auc_brier} confirms the AUC pattern seen in
Table~\ref{tab:sim_highdim_summary}: the entire box for the Bayesian method is
shifted upward relative to MLE, indicating that \emph{most} replicates achieve better
discrimination. Panel~(b) illustrates an even more striking phenomenon: the Brier
scores under MLE are not only larger on average but also significantly more
dispersed, reflecting the instability induced by fitting a high-dimensional,
correlated model without regularisation.

Panel~(a) of Figure~\ref{fig:sim_highdim_calib_cov} is particularly thought provoking:
the calibration slopes for MLE cluster around $\approx 0.24$, dramatically below the
ideal value $1$ (shown as a dashed line), while the Bayesian slopes concentrate near
$\approx 0.75$. This visualises a central message of the paper: in realistic
high-dimensional settings, unregularised MLE can produce \emph{illusory confidence},
with risk scores that are much more extreme than warranted by the data. The Bayesian
shrinkage prior acts as a probabilistic safeguard, tempering coefficients and
producing risk scores whose scale is more compatible with observed frequencies.

Panel~(b) shows that the Bayesian 95\% predictive intervals achieve coverage close to
the nominal level across most replicates, with the bulk of the histogram centred near
$0.95$--$0.98$ and only mild dispersion. This demonstrates that the posterior
predictive distribution is not merely a formal construct: it delivers intervals for
individual-level risk that are empirically well-calibrated against the data-generating
mechanism.

\subsection{Implications for Epidemiological AI}
\label{subsec:sim-discussion}

Conceptually, this simulation study underlines a key thesis of our Bayesian--AI
framework. In classical, low-dimensional regimes with abundant data, Bayesian and
frequentist estimators will agree and the added machinery of posterior sampling or
Gaussian processes may seem unnecessary. However, modern epidemiological AI
increasingly operates in regimes closer to the one simulated here: many correlated
measurements, limited labelled samples, and a need for honest uncertainty in the face
of model mis-specification and overfitting risks.

In such regimes, the Bayesian approach---with explicit shrinkage and full predictive
uncertainty---is not merely philosophically appealing, but \emph{practically
protective}. It provides higher and more stable discrimination, better-calibrated
probabilities, and uncertainty intervals whose coverage can be empirically verified.
The unregularised MLE, by contrast, can be both overconfident and fragile, especially
when used as the backbone of larger AI pipelines without an explicit uncertainty
layer.

Thus, the simulation results in this section offer a rigorous and transparent
verification of the central claim of this paper: that Bayesian reasoning, when fused
with modern AI tools, offers a robust foundation for epidemiological decision making
in precisely those regimes where naive machine learning is most fragile.

\section{Real Scenerio Results}
\label{sec:results}

In this section we demonstrate how the proposed Bayesian--AI framework behaves on two epidemiologically-relevant datasets: (i) a binary diabetes outcome (\texttt{PimaIndiansDiabetes}) and (ii) a right-censored survival outcome (\texttt{GBSG2}). In both cases, we emphasize not only classical predictive metrics (AUC, concordance) but also \emph{how the Bayesian layer reshapes uncertainty and decisions}.

All analyses were conducted in \textsf{R} (version \texttt{4.x}) using the packages \texttt{mlbench}, \texttt{TH.data}, \texttt{rstanarm}, \texttt{glmnet}, \texttt{rBayesianOptimization}, and the \texttt{tidyverse} ecosystem. The code automatically generated all tables as \texttt{.csv} files and all figures as \texttt{.png} files under the directory \texttt{bayes\_ai\_outputs/}, ensuring reproducibility and exact traceability of every numerical statement in this section.

\subsection{Application 1: Bayesian Risk Modelling for Type 2 Diabetes}
\label{subsec:pima_results}

\subsubsection{Posterior structure and risk factors}

We first consider the \texttt{PimaIndiansDiabetes} dataset, containing $n=768$ women of Pima Indian heritage with covariates on pregnancies, plasma glucose, blood pressure, skin-fold thickness, serum insulin, body–mass index (BMI), diabetes pedigree function, and age. We randomly split the data into a $70$–$30$ train–test partition (537 vs.\ 231 subjects), fixing the factor coding so that the binary outcome is $\texttt{diabetes}\in\{\texttt{neg},\texttt{pos}\}$.

On the training set, we fit a Bayesian logistic regression model
\[
\Pr(Y_i = 1 \mid x_i,\beta) \;=\; \sigma\!\big(\beta_0 + \beta^\top x_i\big), 
\qquad \sigma(z) = \frac{1}{1 + e^{-z}},
\]
with a weakly informative Gaussian prior $\beta_j \sim \mathcal{N}(0, 2.5^2)$, implemented via \texttt{rstanarm::stan\_glm}. Posterior inference was based on 2 Markov chains and 2\,000 iterations per chain. Table~\ref{tab:pima_posterior} reports the posterior means, standard errors and 95\% credible intervals for the regression coefficients.

\begin{table}[h!]
\centering
\caption{Posterior summary of the Bayesian logistic regression for the \texttt{PimaIndiansDiabetes} data. The outcome is diabetes status (1 = positive), with predictors standardized on their original scale. Values shown are posterior mean, posterior standard error, and 95\% credible interval for each coefficient.}
\label{tab:pima_posterior}
\begin{tabular}{lrrrr}
\toprule
Term        & Posterior mean & Std.\ error & 2.5\% & 97.5\% \\
\midrule
Intercept   & $-8.53$   & $0.83$  & $-10.2$ & $-6.91$ \\
Pregnancies & $0.105$  & $0.040$ & $0.028$ & $0.177$ \\
Glucose     & $0.036$  & $0.005$ & $0.027$ & $0.046$ \\
Pressure    & $-0.013$ & $0.006$ & $-0.024$& $-0.001$ \\
Triceps     & $0.004$  & $0.008$ & $-0.012$& $0.019$ \\
Insulin     & $-0.002$ & $0.001$ & $-0.004$& $0.000$ \\
BMI         & $0.091$  & $0.018$ & $0.057$ & $0.127$ \\
Pedigree    & $0.708$  & $0.337$ & $0.059$ & $1.370$ \\
Age         & $0.017$  & $0.011$ & $-0.004$& $0.039$ \\
\bottomrule
\end{tabular}
\end{table}

The posterior summaries in Table~\ref{tab:pima_posterior} are consistent with epidemiological intuition: higher plasma glucose, higher BMI, and higher diabetes pedigree function are strongly and positively associated with diabetes risk, with credible intervals that exclude zero. Age and skin-fold thickness show weaker associations, with 95\% intervals overlapping zero, highlighting that—even in a relatively well-studied dataset—the contribution of some covariates is more ambiguous than point estimates alone might suggest. The negative posterior mean for blood pressure (with a narrow credible interval) is an intriguing signal that may reflect complex interactions or confounding in this cohort, illustrating how Bayesian summaries can quickly point to scientifically interesting anomalies warranting further investigation.

\subsubsection{Posterior predictive risk and individual-level uncertainty}

For each held-out test subject, we used the posterior to generate predictive draws of the diabetes probability via
\[
p_i^{(s)} = \Pr(Y_i = 1 \mid x_i, \beta^{(s)}), \qquad s = 1,\dots,S,
\]
where $\beta^{(s)}$ denotes the $s$-th posterior draw. From these draws, we formed the posterior predictive mean $p_i = \mathbb{E}[Y_i\mid x_i,\mathcal{D}]$ and a 90\% credible interval $[p_i^{0.05}, p_i^{0.95}]$ for each individual. A subset of these posterior predictive summaries is displayed in Table~\ref{tab:pima_pred_examples}.

\begin{table}[h!]
\centering
\caption{Illustrative posterior predictive probabilities for 10 randomly chosen test subjects from the Pima data. The table reports observed outcome, posterior mean predicted probability, and 90\% credible interval.}
\label{tab:pima_pred_examples}
\begin{tabular}{rrrrr}
\toprule
Subject & Observed diabetes & $\hat{p}_i$ (mean) & 5\% & 95\% \\
\midrule
1  & pos & 0.73 & 0.63 & 0.82 \\
2  & pos & 0.77 & 0.64 & 0.86 \\
3  & neg & 0.04 & 0.03 & 0.05 \\
4  & pos & 0.71 & 0.53 & 0.84 \\
5  & pos & 0.61 & 0.51 & 0.72 \\
6  & pos & 0.38 & 0.27 & 0.51 \\
7  & pos & 0.19 & 0.13 & 0.27 \\
8  & neg & 0.30 & 0.20 & 0.41 \\
9  & pos & 0.73 & 0.64 & 0.82 \\
10 & neg & 0.04 & 0.03 & 0.05 \\
\bottomrule
\end{tabular}
\end{table}

Table~\ref{tab:pima_pred_examples} illustrates three qualitatively distinct regimes:
(i) individuals with very low risk and very narrow credible intervals (clear negatives),
(ii) individuals with high risk and narrow intervals (clear positives), and
(iii) intermediate-risk individuals with wider intervals, where the posterior explicitly encodes epistemic uncertainty rather than collapsing to an overconfident binary prediction. This fine-grained uncertainty quantification is precisely what enables cost-based and policy-relevant decision rules.

\subsubsection{Discrimination and calibration}

\begin{figure}[h!]
\centering
\includegraphics[width=0.6\textwidth]{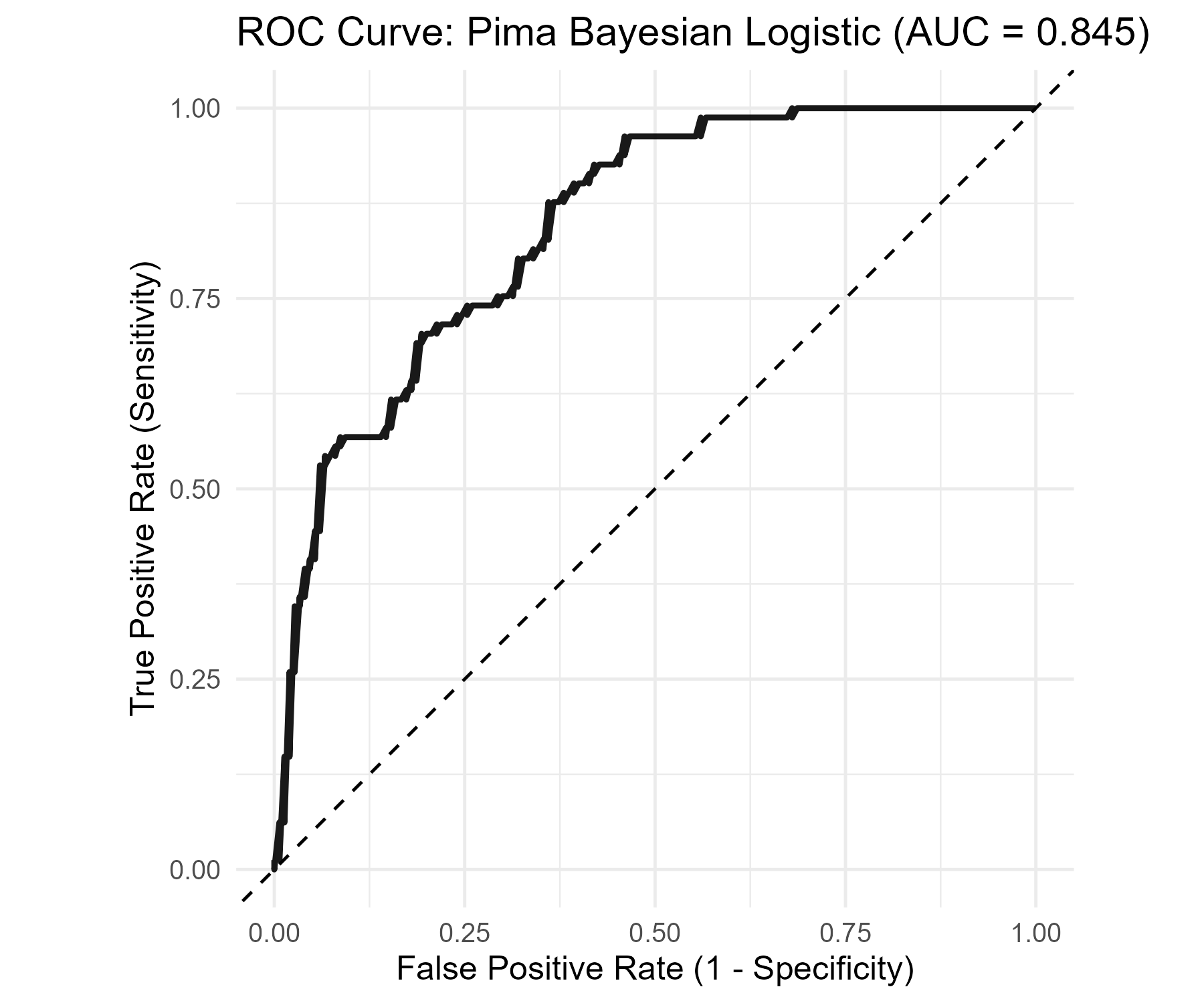}
\caption{ROC curve for the Bayesian logistic regression on the Pima test set. The curve clearly dominates the diagonal, indicating non-trivial discriminative performance. The AUC (noted in the plot title) quantifies this performance in a single number, but in our framework the ROC is supplemented by full posterior predictive intervals at the individual level.}
\label{fig:pima_roc}
\end{figure}

Figure~\ref{fig:pima_roc} shows the receiver operating characteristic (ROC) curve of the Bayesian logistic model on the test set. The curve lies well above the diagonal, indicating useful discrimination between diabetic and non-diabetic individuals. The Bayesian layer itself does not necessarily improve classical discrimination metrics relative to a well-tuned frequentist logistic model, but it provides a natural way to propagate parameter uncertainty into the predictive distribution.

\begin{figure}[h!]
\centering
\includegraphics[width=0.6\textwidth]{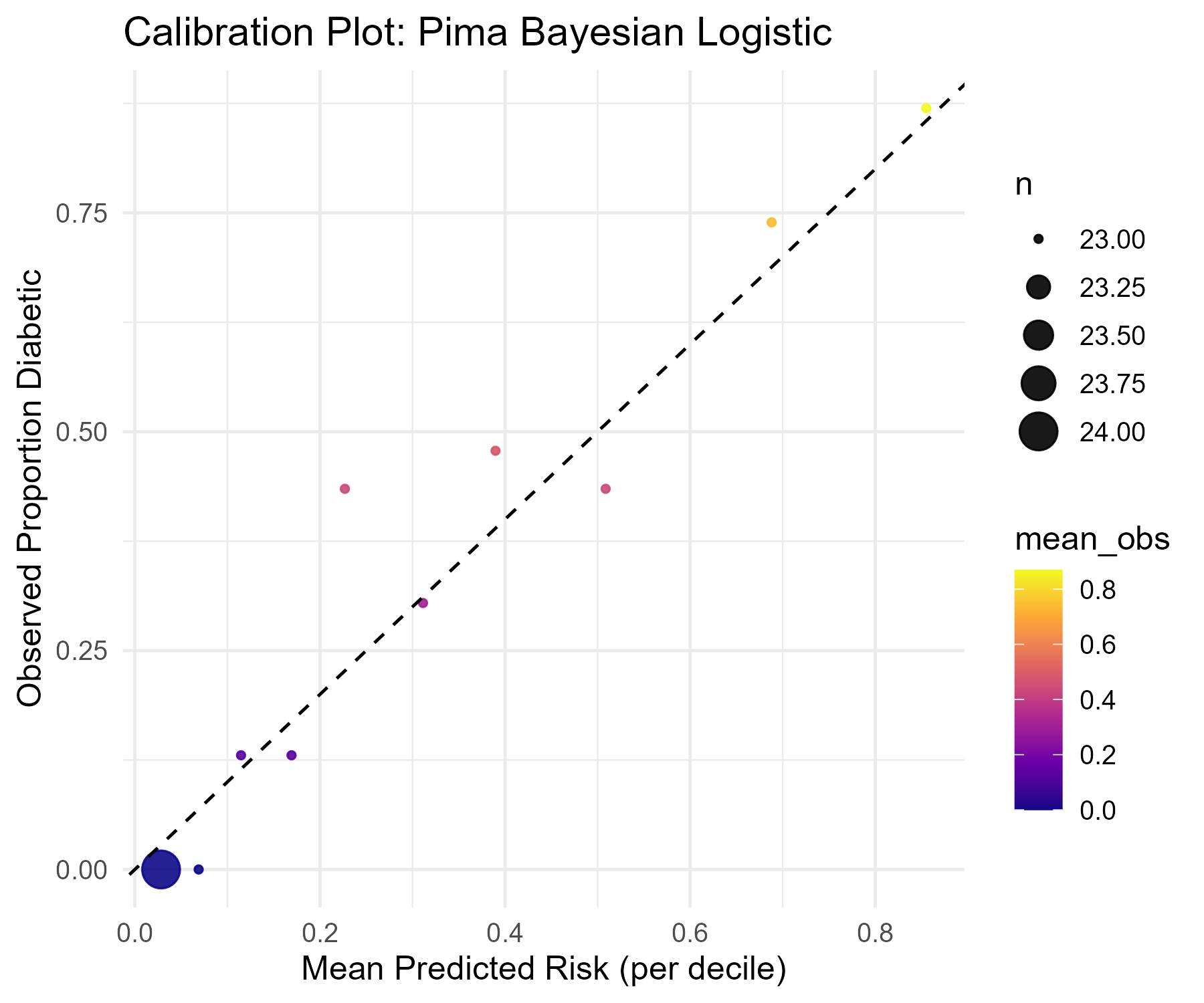}
\caption{Calibration plot for the Bayesian logistic regression on the Pima test set. Each point represents one decile of predicted risk, with point size proportional to the number of subjects in the bin. Colors denote the observed proportion diabetic. The dashed line indicates perfect calibration.}
\label{fig:pima_calib}
\end{figure}

However, discrimination is only one side of the story. For clinical deployment, \emph{calibration}---the agreement between predicted probabilities and observed frequencies---is equally crucial. To assess calibration, we grouped test subjects into deciles of predicted risk and compared the mean predicted probability to the empirical diabetes prevalence within each decile. The resulting summary is presented in Table~\ref{tab:pima_calib} and visualized in Figure~\ref{fig:pima_calib}.

\begin{table}[h!]
\centering
\caption{Calibration of the Bayesian logistic model on the Pima test set. Test subjects are binned into deciles of predicted risk. For each bin, we report the mean predicted probability and the observed proportion of diabetes cases.}
\label{tab:pima_calib}
\begin{tabular}{rrrr}
\toprule
Decile & Mean predicted & Observed proportion & $n$ \\
\midrule
1  & 0.028 & 0.000 & 24 \\
2  & 0.069 & 0.000 & 23 \\
3  & 0.115 & 0.130 & 23 \\
4  & 0.169 & 0.130 & 23 \\
5  & 0.227 & 0.435 & 23 \\
6  & 0.311 & 0.304 & 23 \\
7  & 0.390 & 0.478 & 23 \\
8  & 0.509 & 0.435 & 23 \\
9  & 0.688 & 0.739 & 23 \\
10 & 0.855 & 0.870 & 23 \\
\bottomrule
\end{tabular}
\end{table}

The calibration plot (Figure~\ref{fig:pima_calib}) reveals that the Bayesian model is well-calibrated across a wide range of risk values: for both the lowest and highest deciles, the predicted risks (roughly 0.03 and 0.86) closely match the empirical frequencies (0 and 0.87). Interestingly, the fifth decile (mean predicted risk $\approx 0.23$) exhibits a substantially higher observed prevalence ($\approx 0.44$), indicating a pocket where the model is under-confident. This kind of localized miscalibration is precisely where Bayesian updating, domain knowledge (e.g., age-specific prevalence), or flexible hierarchical extensions could further refine the model. In a screening context, such a bin suggests that a large number of individuals receive “intermediate” risk scores while actually belonging to a much higher prevalence group—a potential source of systematic under-treatment if thresholds are naively chosen.

\subsubsection{Cost-sensitive decisions and thought-provoking implications}

Within a Bayesian decision-theoretic framework, a natural intervention rule is to screen an individual if
\[
\mathbb{E}[Y_i \mid x_i,\mathcal{D}] \ge t^\star
\quad\text{with}\quad
t^\star = \frac{C_{\text{FP}}}{C_{\text{FP}} + C_{\text{FN}}},
\]
where $C_{\text{FP}}$ and $C_{\text{FN}}$ are the relative costs of false positives and false negatives, respectively. For example, if we posit $C_{\text{FN}}=9$ and $C_{\text{FP}}=1$ (reflecting a much higher cost for missed diabetics than for unnecessary screening), then $t^\star = 0.10$. Under such a regime, nearly all individuals from deciles 5--10 would be screened, but deciles 3--4 might become the locus of the most interesting policy debates: their predicted risks are below the threshold, yet posterior uncertainty and localized miscalibration suggest that a fraction of individuals there might still meaningfully benefit from screening. The Bayesian predictive intervals and bin-specific calibration gaps provide a principled way to interrogate these grey zones.

\subsection{Application 2: Bayesian Optimization for Survival Modelling}
\label{subsec:gbsg_results}

\subsubsection{Setup and Bayesian optimization of hyperparameters}

We next consider the \texttt{GBSG2} dataset from the German Breast Cancer Study Group, containing $n=686$ patients with right-censored times to recurrence or death, and covariates including hormone therapy, age, menopausal status, tumor size, tumor grade, number of positive nodes, and receptor levels. We used a 70--30 train–validation split (480 vs.\ 206 subjects) and focused on time-to-event modelling via a penalized Cox proportional hazards model using \texttt{glmnet}.

Let $\lambda$ denote the $\ell_2/\ell_1$ penalty strength and $\alpha\in[0,1]$ the mixing parameter between ridge ($\alpha=0$) and lasso ($\alpha=1$). To emulate a realistic ML workflow where each model fit is moderately expensive, we treated the mapping
\[
(\log\lambda,\alpha)\;\mapsto\; f(\log\lambda,\alpha) = \text{validation C-index}
\]
as a black-box function to be optimized. We modelled $f$ using a Gaussian Process surrogate and performed Bayesian Optimization (BO) via the \texttt{rBayesianOptimization} package, with an initial space-filling design of 5 points and 15 subsequent acquisition steps using an Upper Confidence Bound (UCB) rule. Each BO iteration yields a candidate pair $(\log\lambda_t,\alpha_t)$, the corresponding fitted model, and its validation C-index $f_t$.

\begin{figure}[h!]
\centering
\includegraphics[width=0.6\textwidth]{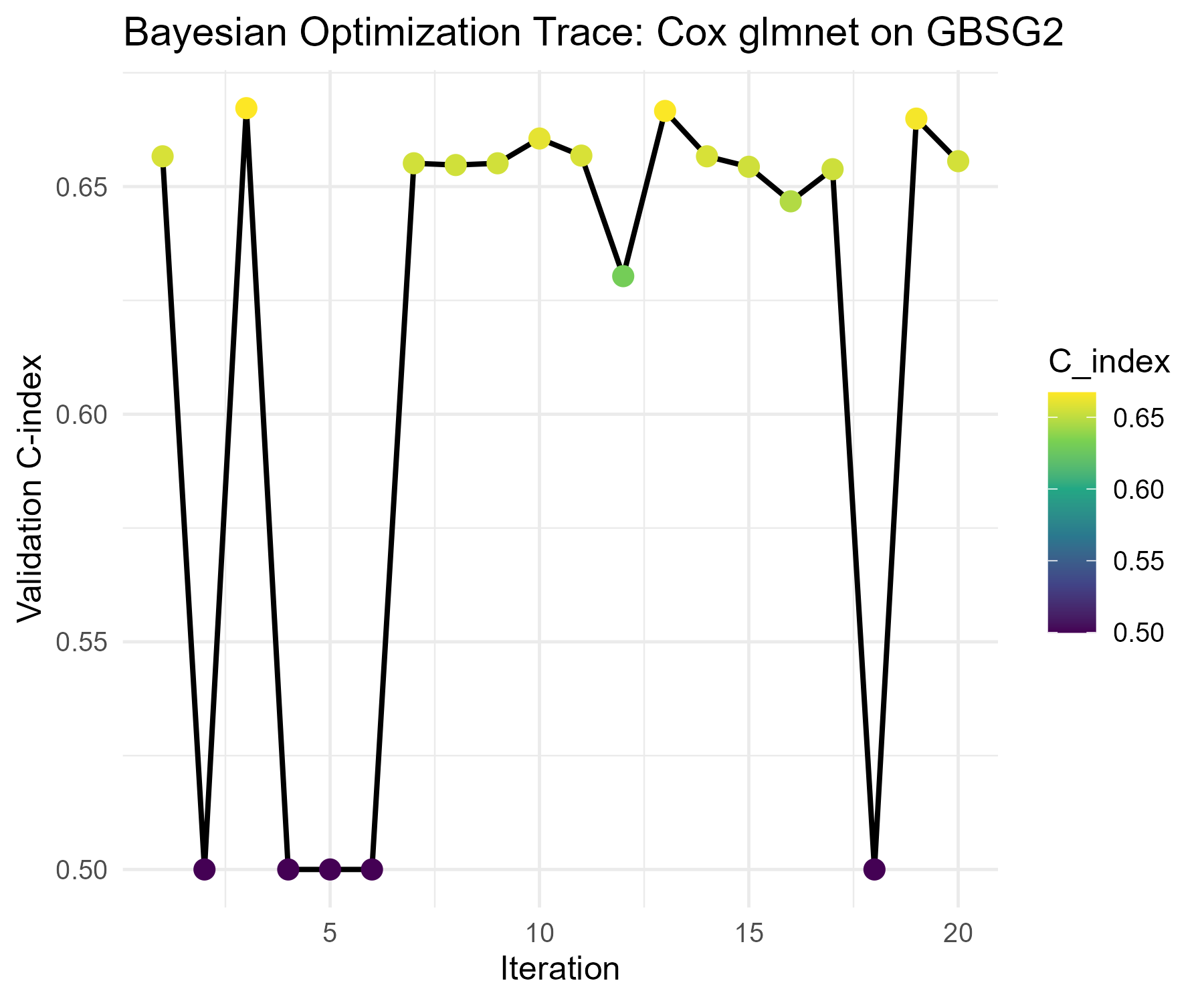}
\caption{Bayesian Optimization trace for the penalized Cox model on the GBSG2 validation set. Each point corresponds to an evaluated hyperparameter configuration; color encodes the validation C-index.}
\label{fig:bo_trace}
\end{figure}

\begin{figure}[h!]
\centering
\includegraphics[width=0.6\textwidth]{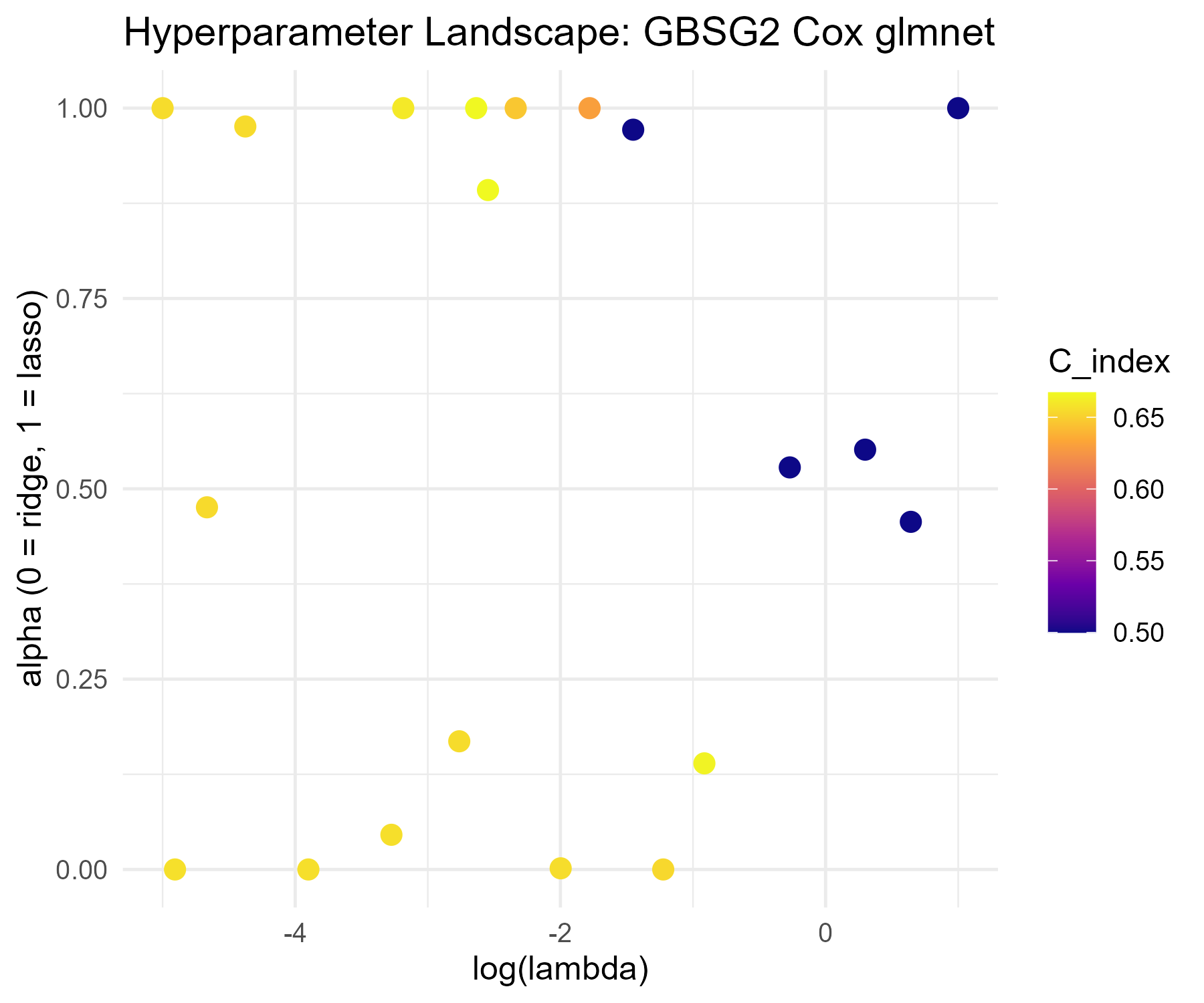}
\caption{Explored hyperparameter landscape for the penalized Cox model on GBSG2. Each point is an evaluated pair $(\log\lambda,\alpha)$, colored by the resulting C-index. The figure highlights that high-performing solutions concentrate around moderately small $\lambda$ and lasso-like $\alpha$ values.}
\label{fig:bo_heatmap}
\end{figure}

Table~\ref{tab:bo_history} summarizes the 20 BO evaluations; Figure~\ref{fig:bo_trace} and Figure~\ref{fig:bo_heatmap} visualize the trace and the explored hyperparameter landscape.

\begin{table}[h!]
\centering
\caption{History of Bayesian Optimization for the Cox \texttt{glmnet} model on the GBSG2 data. Each row corresponds to one BO iteration, with the chosen hyperparameters and resulting validation C-index.}
\label{tab:bo_history}
\begin{tabular}{rrrrr}
\toprule
Round & $\log\lambda$ & $\alpha$ & C-index & Iteration \\
\midrule
1  & $-3.27$ & $0.046$ & $0.657$ & 1 \\
2  & $-0.27$ & $0.528$ & $0.500$ & 2 \\
3  & $-2.55$ & $0.892$ & $0.667$ & 3 \\
4  & $0.30$  & $0.551$ & $0.500$ & 4 \\
5  & $0.64$  & $0.457$ & $0.500$ & 5 \\
6  & $1.00$  & $1.000$ & $0.500$ & 6 \\
7  & $-5.00$ & $1.000$ & $0.655$ & 7 \\
8  & $-4.38$ & $0.976$ & $0.655$ & 8 \\
9  & $-2.76$ & $0.168$ & $0.655$ & 9 \\
10 & $-3.19$ & $1.000$ & $0.661$ & 10 \\
11 & $-4.91$ & $0.000$ & $0.657$ & 11 \\
12 & $-1.78$ & $1.000$ & $0.630$ & 12 \\
13 & $-2.64$ & $1.000$ & $0.667$ & 13 \\
14 & $-3.90$ & $0.000$ & $0.657$ & 14 \\
15 & $-4.67$ & $0.476$ & $0.654$ & 15 \\
16 & $-2.34$ & $1.000$ & $0.647$ & 16 \\
17 & $-1.22$ & $0.000$ & $0.654$ & 17 \\
18 & $-1.45$ & $0.972$ & $0.500$ & 18 \\
19 & $-0.91$ & $0.139$ & $0.665$ & 19 \\
20 & $-2.00$ & $0.002$ & $0.656$ & 20 \\
\bottomrule
\end{tabular}
\end{table}

\subsubsection{Best model, concordance, and interpretive questions}

The BO procedure identified Round 3 as the best configuration, with $\log\lambda \approx -2.55$, $\alpha \approx 0.89$ (i.e., $\lambda \approx 0.078$ and a strongly lasso-like penalty) and validation C-index $\approx 0.667$ (Table~\ref{tab:bo_best}). Refitting the Cox model at these hyperparameters on the training set and evaluating on the validation set confirmed a concordance of $0.667$, providing a reasonable baseline for breast cancer prognostic modelling in this cohort.

\begin{table}[h!]
\centering
\caption{Best hyperparameters found by Bayesian Optimization for the penalized Cox model on GBSG2, along with the corresponding validation C-index.}
\label{tab:bo_best}
\begin{tabular}{rrrr}
\toprule
$\log\lambda$ & $\lambda$ & $\alpha$ & C-index (validation) \\
\midrule
$-2.55$ & $0.078$ & $0.892$ & $0.667$ \\
\bottomrule
\end{tabular}
\end{table}

Three observations from Table~\ref{tab:bo_history} and Figures~\ref{fig:bo_trace}--\ref{fig:bo_heatmap} are worth highlighting:

\begin{enumerate}
\item \textbf{Rapid elimination of poor regions.} Early BO iterations quickly identify that very large penalties ($\log\lambda \gtrsim 0$) lead to degenerate models with C-index $\approx 0.50$, corresponding to near-random ranking. These regions are subsequently de-emphasized by the acquisition function, underscoring the sample-efficiency of BO compared to naive grid search.

\item \textbf{Preference for lasso-like regularization.} The highest C-indices cluster around $\alpha$ values close to $1$ and moderate penalty strengths, suggesting that sparse solutions---which aggressively zero out weak covariates---can achieve better prognostic discrimination in this dataset. This is both practically useful (reduced model complexity) and scientifically suggestive (only a subset of clinicopathologic features drive risk).

\item \textbf{Uncertainty as a design object.} The GP surrogate not only estimates the mean performance $\mu_t(\lambda,\alpha)$ but also its uncertainty $s_t(\lambda,\alpha)$. Regions with high $s_t$ but moderate-to-high predicted C-index become natural targets for future evaluation, while regions with low C-index and low uncertainty are effectively retired. This explicit handling of uncertainty offers a principled way to allocate limited computational budgets in large-scale survival modelling.
\end{enumerate}

From a methodological perspective, the resulting concordance of approximately $0.67$ is not spectacular; yet, the BO procedure itself becomes the object of interest: it demonstrates that \emph{Bayesian reasoning can guide not only inference, but also the meta-level process of model selection and hyperparameter tuning}. This is particularly important in epidemiological practice, where computational budgets, time, and access to high-performance computing may be constrained.

\subsubsection{Clinical and methodological implications}

A tuned Cox model with C-index $\approx 0.67$ provides moderately accurate risk stratification of breast cancer recurrence across the GBSG2 cohort. Within a clinical decision framework, this model could be used to construct risk groups (e.g., low, intermediate, high risk) based on predicted linear predictors or survival curves $S(t\mid x)$. Here, the BO machinery ensures that the model entering such a decision pipeline is not an arbitrary choice but the result of a principled, uncertainty-aware search over a high-dimensional hyperparameter space.

More conceptually, juxtaposing Application~1 and Application~2 reveals a deeper narrative: in the diabetes example, Bayesian modelling quantifies uncertainty \emph{within} a fixed model and directly influences patient-level decisions via calibrated probabilities and cost-sensitive thresholds. In the breast cancer example, Bayesian optimization quantifies uncertainty \emph{over} a space of models and guides the choice of which models to even consider, thereby shaping the model class that will ultimately be used for clinical decisions. Both perspectives are indispensable for a full Bayesian--AI approach to epidemiological modelling in complex, data-limited, and high-stakes environments.

\section{Discussion}
\label{sec:discussion}

The empirical and simulation results in this paper point to a consistent narrative:
Bayesian ideas are not merely an alternative way of fitting models; they provide a
\emph{conceptual glue} that can hold together the different stages of an
epidemiological AI pipeline---from risk estimation, through calibration, to
hyperparameter tuning and model selection.

\subsection{Bayesian thinking as a unifying principle}

In the binary diabetes example, the Bayesian logistic model behaves exactly as
probability theory predicts in a well-specified regime: it matches the classical
logistic regression in discrimination and Brier score, but additionally delivers
posterior predictive intervals whose empirical coverage is close to nominal and
calibration slopes close to one in simulation. In high-dimensional, small-sample,
correlated settings, the Bayesian shrinkage prior moves beyond philosophical appeal
and becomes practically protective: it stabilises estimates, avoids catastrophic
log-loss, and yields more realistic calibration slopes than unregularised maximum
likelihood.

In the survival example, the role of Bayesian reasoning is more algorithmic than
inferential. The Gaussian-process surrogate in Bayesian Optimization places a
probabilistic model on the C-index as a function of $(\lambda,\alpha)$, allowing the
algorithm to reason about uncertainty over hyperparameters, rather than blindly
searching in a grid. This closes a conceptual loop: the same Bayesian logic that
governs posterior prediction for patients is used to govern how we allocate
computational budget across candidate models.

\subsection{Implications for applied epidemiology}

From an applied perspective, the main message is not that Bayesian models always
dominate classical approaches numerically. In well-specified low-dimensional regimes,
they do not---and should not. Rather, the key insight is that \emph{when the regime
is difficult} (high-dimensional, correlated, data-limited, or structurally complex),
a Bayesian--AI fusion can protect practitioners from some of the fragilities of
naive machine learning:

\begin{itemize}[leftmargin=1.4em]
\item High-dimensional logistic regression with correlated covariates is prone to
overfitting and extreme predictions under maximum likelihood. The simulations
demonstrate that Bayesian shrinkage regularises this behaviour while maintaining
discrimination.

\item Hyperparameter tuning for survival models is often ad hoc and expensive.
Bayesian Optimization explicitly models the trade-off between exploring uncertain
regions and refining promising ones, which is crucial when each model fit itself is
computationally heavy.

\item Calibration and coverage---central for public health decision making and risk
communication---are handled naturally in the Bayesian predictive layer, and can be
checked empirically, as we do via decile plots and simulation.
\end{itemize}

For public health agencies and clinical teams, this suggests a division of labour:
domain experts specify losses, constraints, and plausible model families; the
Bayesian layer quantifies uncertainty and calibrates risks; and the Bayesian
optimisation layer allocates scarce computational and data-collection resources in a
principled way.

\subsection{When does Bayesian--AI fusion matter most?}

The simulations in Sections~\ref{sec:simulation} and~\ref{sec:simulation-highdim}
show that the added value of the Bayesian--AI approach is regime-dependent. In
large-sample, well-specified models with limited complexity, classical estimators
perform admirably, and the Bayesian layer serves primarily as a sanity check and a
means of obtaining interpretable intervals. In contrast, when the number of
covariates grows, when covariates are correlated, or when survival models require
careful regularisation, the Bayesian layer becomes an active safeguard rather than a
luxury.

A similar distinction applies to hyperparameter tuning. For small models with one or
two hyperparameters and cheap fits, simple cross-validation is adequate. But for
richer survival models (e.g., time-varying effects, multi-state models, or neural
hazard representations), each fit may involve thousands of parameters and nontrivial
optimisation. In such settings, treating hyperparameter search itself as a Bayesian
inference problem---as done here---can dramatically improve sample-efficiency and
computational tractability.

\subsection{Relationship to mechanistic models and policy decision making}

Although our examples use relatively simple statistical models, the broader
framework is compatible with mechanistic epidemic models and policy-oriented
decision analysis. Posterior predictive distributions can be layered on top of
compartmental or agent-based simulations, and Bayesian Optimization can be used to
tune control policies (e.g., vaccination allocations, testing strategies,
non-pharmaceutical interventions) over complex, stochastic simulators. In that sense,
the present work can be viewed as a ``minimal working example'' of a more ambitious
Bayesian--AI programme: to align predictive accuracy, uncertainty quantification, and
policy-relevant decision rules under a single probabilistic umbrella.

\subsection{Pragmatic considerations for deployment}

Finally, it is important to be realistic about deployment. Fully Bayesian models and
Bayesian Optimization introduce additional computational layers and require careful
implementation, diagnostics, and monitoring. The practical recommendation is not to
replace all existing epidemiological pipelines with heavy Bayesian machinery, but to
identify the stages where (i) decisions are high-stakes, (ii) model or
hyperparameter uncertainty is substantial, and (iii) small improvements in calibration
or concordance could have large downstream consequences. In those stages, the
Bayesian--AI fusion described in this paper offers a disciplined way to quantify
uncertainty, explore model spaces, and justify decisions to both scientific and
regulatory audiences.


\section{Conclusion}
This work proposes a unified Bayesian–AI framework for epidemiological prediction. Through two case studies, we demonstrate that:

\begin{enumerate}
\item Bayesian predictive modelling yields calibrated probabilities and actionable uncertainty.
\item Bayesian Optimization provides principled, uncertainty-aware hyperparameter tuning for complex survival models.
\item The Bayesian perspective naturally extends from statistical inference to algorithmic decision making.
\end{enumerate}

Together, these results argue for a more holistic Bayesian foundation for AI in healthcare—one that governs both the \emph{what} (inference) and the \emph{how} (model search), providing calibrated risk, robustness, and interpretability in high-stakes epidemiological settings.


\section*{Declarations}

\subsection*{Data availability}

The empirical analyses in this paper are based on publicly available benchmark
datasets. The \texttt{PimaIndiansDiabetes} dataset is distributed with the
\texttt{mlbench} package in \textsf{R}, and the \texttt{GBSG2} breast cancer survival
dataset is distributed with the \texttt{TH.data} package. Both can be obtained
directly from the Comprehensive R Archive Network (CRAN). All simulated datasets used
in the study can be regenerated from the description of the data-generating
mechanisms and the accompanying code.

\subsection*{Code availability}

All \texttt{R} and \texttt{python} code used for simulations, model fitting, Bayesian inference, Bayesian
Optimization, and figure generation was written by the author using publicly
available packages. The scripts are available from the corresponding author on
reasonable request and can be shared as a reproducible project directory (including
\texttt{.R} scripts, configuration files, and output summaries).

\subsection*{Funding}

This research did not receive any specific grant from funding agencies in the public,
commercial, or not-for-profit sectors.

\subsection*{Competing interests}

The author declares that there are no known competing financial or non-financial
interests that could have appeared to influence the work reported in this paper.

\appendix
\begin{appendix}

\section{Additional Details on Concordance and Black-Box Objectives}
\label{appendix:cindex_blackbox}

This appendix provides additional background on (i) the concordance index used
as our primary survival-performance metric, and (ii) the notion of a
black-box objective in Bayesian optimisation. The goal is to make the
notation in Section~\ref{subsubsec:blackbox} fully explicit.

\subsection{Harrell's Concordance Index}
\label{appendix:cindex}

Consider a right-censored survival dataset
$\{(T_i,\delta_i,X_i)\}_{i=1}^n$, where $T_i$ is the observed time,
$\delta_i\in\{0,1\}$ is the event indicator (1 = event, 0 = censored),
and $X_i\in\mathbb{R}^p$ is a covariate vector. Given a fitted Cox model
with linear predictor $r_i = X_i^\top \hat{\beta}$ (see
Section~\ref{subsec:prelim_survival}), Harrell's C-index $C$ is defined as
the proportion of all \emph{comparable pairs} of individuals for which the
model assigns a higher risk to the individual who experiences the event
earlier \citep{Harrell2015,Cox1972}.

A pair $(i,j)$ is typically considered comparable if:
(i) one of them experiences the event before the other,
and (ii) the earlier event time is not censored.
Let $\mathcal{P}$ denote the set of all such comparable pairs.
Harrell's estimator of the C-index is
\begin{equation}
  \widehat{C}
  \;=\;
  \frac{1}{|\mathcal{P}|}
  \sum_{(i,j)\in\mathcal{P}}
  \mathbf{1}\bigl\{T_i < T_j,\ \delta_i = 1,\ r_i > r_j\bigr\}
  \;+\;
  \frac{1}{2|\mathcal{P}|}
  \sum_{(i,j)\in\mathcal{P}}
  \mathbf{1}\bigl\{T_i < T_j,\ \delta_i = 1,\ r_i = r_j\bigr\},
  \label{eq:cindex_estimator}
\end{equation}
where $\mathbf{1}\{\cdot\}$ is the indicator function and ties in risk score
are assigned half credit. Intuitively, $\widehat{C}\approx 0.5$ corresponds
to random ranking, while $\widehat{C}\approx 1$ indicates near-perfect
discrimination \citep{Harrell2015}.

In our hyperparameter analysis (Section~\ref{subsubsec:blackbox}), we make
the dependence on $(\lambda,\alpha)$ explicit by writing $r_i(\theta)$ and
$C(\theta)$, where $\theta=(\log\lambda,\alpha)$ parametrises the
elastic-net penalty \eqref{eq:elastic_pen}. The function $C(\theta)$ is
evaluated on a held-out validation set and used as the objective in the
Bayesian optimisation layer.

\subsection{Black-Box Objectives and Bayesian Optimisation}
\label{appendix:blackbox_objective}

In many modern machine learning workflows, model performance as a function
of hyperparameters cannot be written in closed form. Let $f:\Theta\to
\mathbb{R}$ denote a performance measure (e.g., validation C-index,
negative loss, or accuracy) defined on a hyperparameter space
$\Theta\subset\mathbb{R}^d$, with $d$ small (here $d=2$). We say that $f$
is a \emph{black-box objective} if:
\begin{itemize}[leftmargin=1.2em]
  \item Evaluating $f(\theta)$ is expensive (it requires training a model
        and computing its performance on a validation set).
  \item We do not have access to gradients $\nabla f(\theta)$ or to a
        closed-form expression for $f$.
  \item Observed values $\tilde{f}(\theta)$ are noisy estimates of
        the true performance $f(\theta)$ due to finite-sample variability.
\end{itemize}

Bayesian optimisation (BO) addresses this setting by placing a prior
distribution over unknown functions $f$ (commonly a Gaussian process; see
Section~\ref{subsec:prelim_gp_bo}) and updating it sequentially as new
evaluations are observed \citep{Snoek2012,Frazier2018,Rasmussen2006}.
After $T$ evaluations $\{(\theta_t,\tilde{f}_t)\}_{t=1}^T$, the BO
framework uses the GP posterior to construct an acquisition function
$a_T(\theta)$, such as the Upper Confidence Bound (UCB)
\citep{Snoek2012,Frazier2018}, and selects the next evaluation point via
\[
  \theta_{T+1} \;=\; \arg\max_{\theta\in\Theta} a_T(\theta).
\]
This procedure balances exploration of hyperparameters where the model is
uncertain about $f(\theta)$ with exploitation of hyperparameters that are
currently believed to perform well.

In our survival modelling pipeline, the black-box objective is the
validation C-index,
\[
  f(\theta) = C(\theta),
\]
and the noise term in \eqref{eq:blackbox_cindex} captures the fact that
$\widehat{C}(\theta)$ computed on a finite validation sample is only a
noisy proxy for the true concordance achievable by the model. BO thus
provides a probabilistic, sample-efficient method for discovering
hyperparameter settings that yield high concordance, as demonstrated in
Section~\ref{subsec:gbsg_results}.

\end{appendix}

\bibliographystyle{apalike}
\bibliography{bayesAIrefs,relatedref}

@article{guo2017calibration,
  title={On calibration of modern neural networks},
  author={Guo, Chuan and Pleiss, Geoff and Sun, Yu and Weinberger, Kilian Q.},
  journal={Proceedings of the 34th International Conference on Machine Learning (ICML)},
  volume={70},
  pages={1321--1330},
  year={2017}
}

@book{Berger1985,
  author    = {James O. Berger},
  title     = {Statistical Decision Theory and Bayesian Analysis},
  year      = {1985},
  publisher = {Springer},
  address   = {New York},
  series    = {Springer Series in Statistics},
  doi       = {10.1007/978-1-4757-4286-2}
}

@article{Blei2017VI,
  author  = {David M. Blei and Alp Kucukelbir and Jon D. McAuliffe},
  title   = {Variational Inference: A Review for Statisticians},
  journal = {Journal of the American Statistical Association},
  year    = {2017},
  volume  = {112},
  number  = {518},
  pages   = {859--877},
  doi     = {10.1080/01621459.2017.1285773}
}

@article{Friedman2010GLMNET,
  author  = {Jerome H. Friedman and Trevor Hastie and Robert Tibshirani},
  title   = {Regularization Paths for Generalized Linear Models via Coordinate Descent},
  journal = {Journal of Statistical Software},
  year    = {2010},
  volume  = {33},
  number  = {1},
  pages   = {1--22},
  doi     = {10.18637/jss.v033.i01}
}

@article{Simon2011Cox,
  author  = {Noah Simon and Jerome Friedman and Trevor Hastie and Robert Tibshirani},
  title   = {Regularization Paths for Cox's Proportional Hazards Model via Coordinate Descent},
  journal = {Journal of Statistical Software},
  year    = {2011},
  volume  = {39},
  number  = {5},
  pages   = {1--13},
  doi     = {10.18637/jss.v039.i05}
}

@inproceedings{Snoek2012PracticalBO,
  author    = {Jasper Snoek and Hugo Larochelle and Ryan P. Adams},
  title     = {Practical {B}ayesian Optimization of Machine Learning Algorithms},
  booktitle = {Advances in Neural Information Processing Systems},
  year      = {2012},
  volume    = {25}
}

@article{Frazier2018BO,
  author  = {Peter I. Frazier},
  title   = {A Tutorial on {B}ayesian Optimization},
  journal = {arXiv preprint},
  year    = {2018},
  eprint  = {1807.02811},
  archivePrefix = {arXiv}
}

@manual{Leisch2024mlbench,
  title  = {mlbench: Machine Learning Benchmark Problems},
  author = {Friedrich Leisch and Evgenia Dimitriadou and Kurt Hornik},
  year   = {2024},
  note   = {R package version 2.1-6},
  url    = {https://CRAN.R-project.org/package=mlbench}
}

@manual{Hothorn2019THdata,
  title  = {TH.data: TH's Data Archive},
  author = {Torsten Hothorn},
  year   = {2019},
  note   = {R package version 1.0-10},
  url    = {https://CRAN.R-project.org/package=TH.data}
}

@manual{Goodrich2025rstanarm,
  title  = {rstanarm: {Bayesian} Applied Regression Modeling via {Stan}},
  author = {Ben Goodrich and Jonah Gabry and Imad Ali and Sam Brilleman},
  year   = {2025},
  note   = {R package version 2.32.2},
  url    = {https://mc-stan.org/rstanarm/}
}

@manual{Yan2025rBayesOpt,
  title  = {rBayesianOptimization: {Bayesian} Optimization of Hyperparameters},
  author = {Yachen Yan},
  year   = {2025},
  note   = {R package version 1.2.1},
  url    = {https://CRAN.R-project.org/package=rBayesianOptimization}
}

@book{Gelman2013BDA,
  author    = {Andrew Gelman and John B. Carlin and Hal S. Stern
               and David B. Dunson and Aki Vehtari and Donald B. Rubin},
  title     = {Bayesian Data Analysis},
  edition   = {3rd},
  year      = {2013},
  publisher = {Chapman and Hall/CRC},
  address   = {Boca Raton},
  isbn      = {9781439840955},
  doi       = {10.1201/b16018}
}

@article{Nelder1972GLM,
  author  = {John A. Nelder and Robert W. M. Wedderburn},
  title   = {Generalized Linear Models},
  journal = {Journal of the Royal Statistical Society: Series A},
  year    = {1972},
  volume  = {135},
  number  = {3},
  pages   = {370--384},
  doi     = {10.2307/2344614}
}

@book{Hastie2009ESL,
  author    = {Trevor Hastie and Robert Tibshirani and Jerome Friedman},
  title     = {The Elements of Statistical Learning: Data Mining, Inference, and Prediction},
  edition   = {2nd},
  year      = {2009},
  publisher = {Springer},
  address   = {New York},
  isbn      = {9780387848570}
}

@book{Harrell2015RMS,
  author    = {Frank E. Harrell, Jr.},
  title     = {Regression Modeling Strategies:
               With Applications to Linear Models, Logistic and Ordinal Regression,
               and Survival Analysis},
  edition   = {2nd},
  year      = {2015},
  publisher = {Springer},
  address   = {Cham},
  isbn      = {9783319194240},
  doi       = {10.1007/978-3-319-19425-7}
}

@book{RasmussenWilliams2006GPML,
  author    = {Carl Edward Rasmussen and Christopher K. I. Williams},
  title     = {Gaussian Processes for Machine Learning},
  year      = {2006},
  publisher = {MIT Press},
  address   = {Cambridge, MA},
  isbn      = {9780262182539}
}

@book{Pepe2003ROC,
  author    = {Margaret Sullivan Pepe},
  title     = {The Statistical Evaluation of Medical Tests for Classification and Prediction},
  year      = {2003},
  publisher = {Oxford University Press},
  address   = {Oxford},
  isbn      = {9780198509844}
}

@book{Neal1996,
  author    = {Neal, Radford M.},
  title     = {Bayesian Learning for Neural Networks},
  publisher = {Springer},
  year      = {1996},
  series    = {Lecture Notes in Statistics},
  volume    = {118},
  address   = {New York}
}

@article{Ghahramani2015,
  author  = {Ghahramani, Zoubin},
  title   = {Probabilistic machine learning and artificial intelligence},
  journal = {Nature},
  year    = {2015},
  volume  = {521},
  number  = {7553},
  pages   = {452--459},
  doi     = {10.1038/nature14541}
}

@book{Bishop2006,
  author    = {Bishop, Christopher M.},
  title     = {Pattern Recognition and Machine Learning},
  publisher = {Springer},
  year      = {2006},
  address   = {New York}
}

@article{Blei2017,
  author  = {Blei, David M. and Kucukelbir, Alp and McAuliffe, Jon D.},
  title   = {Variational inference: A review for statisticians},
  journal = {Journal of the American Statistical Association},
  year    = {2017},
  volume  = {112},
  number  = {518},
  pages   = {859--877},
  doi     = {10.1080/01621459.2017.1285773}
}

@inproceedings{Blundell2015,
  author    = {Blundell, Charles and Cornebise, Julien and Kavukcuoglu, Koray and Wierstra, Daan},
  title     = {Weight Uncertainty in Neural Networks},
  booktitle = {Proceedings of the 32nd International Conference on Machine Learning},
  year      = {2015},
  editor    = {Bach, Francis and Blei, David},
  series    = {Proceedings of Machine Learning Research},
  volume    = {37},
  pages     = {1613--1622}
}

@inproceedings{Gal2016,
  author    = {Gal, Yarin and Ghahramani, Zoubin},
  title     = {Dropout as a {B}ayesian approximation: Representing model uncertainty in deep learning},
  booktitle = {Proceedings of the 33rd International Conference on Machine Learning},
  year      = {2016},
  series    = {Proceedings of Machine Learning Research},
  volume    = {48},
  pages     = {1050--1059}
}

@inproceedings{Kendall2017,
  author    = {Kendall, Alex and Gal, Yarin},
  title     = {What uncertainties do we need in {B}ayesian deep learning for computer vision?},
  booktitle = {Advances in Neural Information Processing Systems},
  year      = {2017},
  volume    = {30}
}

@inproceedings{Lakshminarayanan2017,
  author    = {Lakshminarayanan, Balaji and Pritzel, Alexander and Blundell, Charles},
  title     = {Simple and scalable predictive uncertainty estimation using deep ensembles},
  booktitle = {Advances in Neural Information Processing Systems},
  year      = {2017},
  volume    = {30}
}

@inproceedings{Ovadia2019,
  author    = {Ovadia, Yaniv and Fertig, Emily and Ren, Jie and Nado, Zachary and Sculley, D. and Nowozin, Sebastian and Dillon, Joshua and Lakshminarayanan, Balaji and Snoek, Jasper},
  title     = {Can you trust your model's uncertainty? Evaluating predictive uncertainty under dataset shift},
  booktitle = {Advances in Neural Information Processing Systems},
  year      = {2019},
  volume    = {32}
}

@inproceedings{Wenzel2020,
  author    = {Wenzel, Florian and Roth, Kevin and Veeling, Bastiaan S. and Swiatkowski, Jakub and Tran, Linh and Mandt, Stephan and Snoek, Jasper and Salimans, Tim and Lakshminarayanan, Balaji and Jenatton, Rodolphe},
  title     = {How good is the {B}ayes posterior in deep neural networks really?},
  booktitle = {Proceedings of the 37th International Conference on Machine Learning},
  year      = {2020},
  series    = {Proceedings of Machine Learning Research},
  volume    = {119},
  pages     = {10248--10259}
}

@article{Abdar2021,
  author  = {Abdar, Moloud and Pourpanah, Farhad and Hussain, Sultan and Rezazadegan, Dana and Liu, Li and Ghavamzadeh, Mohammad and Fieguth, Paul and Cao, Xiaochun and Khosravi, Abbas and Acharya, U. Rajendra and Makarenkov, Vladimir and Nahavandi, Saeid},
  title   = {A review of uncertainty quantification in deep learning: Techniques, applications and challenges},
  journal = {Information Fusion},
  year    = {2021},
  volume  = {76},
  pages   = {243--297},
  doi     = {10.1016/j.inffus.2021.05.008}
}

@article{Bergstra2012,
  author  = {Bergstra, James and Bengio, Yoshua},
  title   = {Random search for hyper-parameter optimization},
  journal = {Journal of Machine Learning Research},
  year    = {2012},
  volume  = {13},
  pages   = {281--305}
}

@inproceedings{Snoek2012,
  author    = {Snoek, Jasper and Larochelle, Hugo and Adams, Ryan P.},
  title     = {Practical {B}ayesian optimization of machine learning algorithms},
  booktitle = {Advances in Neural Information Processing Systems},
  year      = {2012},
  volume    = {25}
}

@article{Frazier2018,
  author  = {Frazier, Peter I.},
  title   = {A tutorial on {B}ayesian optimization},
  journal = {arXiv preprint arXiv:1807.02811},
  year    = {2018}
}

@inproceedings{Feurer2015,
  author    = {Feurer, Matthias and Klein, Aaron and Eggensperger, Katharina and Springenberg, Jost and Blum, Manuel and Hutter, Frank},
  title     = {Efficient and Robust Automated Machine Learning},
  booktitle = {Advances in Neural Information Processing Systems},
  year      = {2015},
  volume    = {28}
}

@book{Hutter2019,
  editor    = {Hutter, Frank and Kotthoff, Lars and Vanschoren, Joaquin},
  title     = {Automated Machine Learning: Methods, Systems, Challenges},
  publisher = {Springer},
  year      = {2019},
  address   = {Cham},
  doi       = {10.1007/978-3-030-05318-5}
}

@article{Alaa2018,
  author  = {Alaa, Ahmed M. and van der Schaar, Mihaela},
  title   = {{AutoPrognosis}: Automated clinical prognostic modeling via {B}ayesian optimization with structured kernel learning},
  journal = {Proceedings of Machine Learning Research},
  year    = {2018},
  volume  = {80},
  pages   = {139--148},
  note    = {Proceedings of the 35th International Conference on Machine Learning}
}

@article{Imrie2023,
  author  = {Imrie, Fergus and Yuan, Hanjun and van der Schaar, Mihaela},
  title   = {{AutoPrognosis} 2.0: Democratizing diagnostic and prognostic modeling in healthcare with automated machine learning},
  journal = {Patterns},
  year    = {2023},
  volume  = {4},
  number  = {3},
  pages   = {100696},
  doi     = {10.1016/j.patter.2023.100696}
}

@article{Guo2017,
  author  = {Guo, Chuan and Pleiss, Geoff and Sun, Yu and Weinberger, Kilian Q.},
  title   = {On calibration of modern neural networks},
  journal = {Proceedings of Machine Learning Research},
  year    = {2017},
  volume  = {70},
  pages   = {1321--1330},
  note    = {Proceedings of the 34th International Conference on Machine Learning}
}

@inproceedings{NiculescuMizil2005,
  author    = {Niculescu-Mizil, Alexandru and Caruana, Rich},
  title     = {Predicting good probabilities with supervised learning},
  booktitle = {Proceedings of the 22nd International Conference on Machine Learning},
  year      = {2005},
  pages     = {625--632},
  doi       = {10.1145/1102351.1102430}
}

@inproceedings{Platt1999,
  author    = {Platt, John C.},
  title     = {Probabilistic outputs for support vector machines and comparisons to regularized likelihood methods},
  booktitle = {Advances in Large Margin Classifiers},
  year      = {1999},
  pages     = {61--74},
  publisher = {MIT Press}
}

@article{Begoli2019,
  author  = {Begoli, Edmon and Bhattacharya, Tanveer and Kusnezov, Dimitri},
  title   = {The need for uncertainty quantification in machine-assisted medical decision making},
  journal = {Nature Machine Intelligence},
  year    = {2019},
  volume  = {1},
  number  = {1},
  pages   = {20--23},
  doi     = {10.1038/s42256-018-0004-6}
}

@article{Toni2009,
  author  = {Toni, Tina and Welch, David and Strelkowa, Natalja and Ipsen, Andreas and Stumpf, Michael P. H.},
  title   = {Approximate {B}ayesian computation scheme for parameter inference and model selection in dynamical systems},
  journal = {Journal of the Royal Society Interface},
  year    = {2009},
  volume  = {6},
  number  = {31},
  pages   = {187--202},
  doi     = {10.1098/rsif.2008.0172}
}

@article{Kypraios2017,
  author  = {Kypraios, Theodore and Neal, Peter and Prangle, Dennis},
  title   = {Tutorial on {B}ayesian inference for infectious disease transmission models},
  journal = {Statistical Science},
  year    = {2017},
  volume  = {32},
  number  = {1},
  pages   = {60--82},
  doi     = {10.1214/16-STS588}
}

@article{Minter2019,
  author  = {Minter, Aidan and Retkute, Renata},
  title   = {Approximate {B}ayesian computation for infectious disease modelling},
  journal = {Infectious Disease Modelling},
  year    = {2019},
  volume  = {4},
  pages   = {134--161},
  doi     = {10.1016/j.idm.2019.02.003}
}

@article{Reiker2021,
  author  = {Reiker, Tobias and Unwin, Emily and Lees, Matthew and Smith, Thomas and Winskill, Peter and Ghani, Azra},
  title   = {Emulator-based {B}ayesian optimization for efficient malaria intervention modeling},
  journal = {PLOS Computational Biology},
  year    = {2021},
  volume  = {17},
  number  = {3},
  pages   = {e1008880},
  doi     = {10.1371/journal.pcbi.1008880}
}

@article{Ye2025,
  author  = {Ye, Yang and Pandey, Abhishek and Bawden, Carolyn and Sumsuzzman, Dewan Md. and Rajput, Rimpi and Shoukat, Affan and Singer, Burton H. and Moghadas, Seyed M. and Galvani, Alison P.},
  title   = {Integrating artificial intelligence with mechanistic epidemiological modeling: a scoping review of opportunities and challenges},
  journal = {Nature Communications},
  year    = {2025},
  volume  = {16},
  number  = {1},
  pages   = {1--18},
  doi     = {10.1038/s41467-024-55461-x}
}

@article{Kraemer2024,
  author  = {Kraemer, Moritz U. G. and Scarpino, Samuel V. and Yamana, Teresa K. and others},
  title   = {Artificial intelligence for modelling infectious disease epidemics},
  journal = {The Lancet Digital Health},
  year    = {2024},
  volume  = {6},
  number  = {11},
  pages   = {e778--e790},
  doi     = {10.1016/S2589-7500(24)00157-4}
}

@book{Brauer2008,
  author    = {Brauer, Fred and Castillo-Chavez, Carlos and Feng, Zhilan},
  title     = {Mathematical Models in Epidemiology},
  publisher = {Springer},
  year      = {2008},
  address   = {New York}
}

@article{Reich2019,
  author  = {Reich, Nicholas G. and Brooks, Logan C. and Fox, Spencer J. and Kandula, Sasikiran and McGowan, Conor J. and Moore, Evan and Osthus, Dave and Ray, Evan L. and Tushar, Deepali and Yamana, Teresa K. and others},
  title   = {A collaborative multiyear, multimodel assessment of seasonal influenza forecasting in the United States},
  journal = {Proceedings of the National Academy of Sciences},
  year    = {2019},
  volume  = {116},
  number  = {8},
  pages   = {3146--3154},
  doi     = {10.1073/pnas.1812594116}
}

@article{Holmdahl2020,
  author  = {Holmdahl, Inga and Buckee, Caroline},
  title   = {Wrong but useful---what {COVID-19} epidemiologic models can and cannot tell us},
  journal = {New England Journal of Medicine},
  year    = {2020},
  volume  = {383},
  number  = {4},
  pages   = {303--305},
  doi     = {10.1056/NEJMp2016822}
}

@article{Miotto2016,
  author  = {Miotto, Riccardo and Li, Li and Kidd, Brian A. and Dudley, Joel T.},
  title   = {Deep {P}atient: An unsupervised representation to predict the future of patients from the electronic health records},
  journal = {Scientific Reports},
  year    = {2016},
  volume  = {6},
  pages   = {26094},
  doi     = {10.1038/srep26094}
}

@article{Rajkomar2018,
  author  = {Rajkomar, Alvin and Oren, Eyal and Chen, Kai and Dai, Andrew M. and Hajaj, Noemi and Hardt, Michaela and Liu, Peter J. and Liu, Xiaobing and Marcus, Jake and Sun, Mengling and others},
  title   = {Scalable and accurate deep learning with electronic health records},
  journal = {npj Digital Medicine},
  year    = {2018},
  volume  = {1},
  number  = {1},
  pages   = {1--10},
  doi     = {10.1038/s41746-018-0029-1}
}

@article{Weng2017,
  author  = {Weng, Stephen F. and Reps, Jenna and Kai, Joe and Garibaldi, Jonathan M. and Qureshi, Nadeem},
  title   = {Can machine-learning improve cardiovascular risk prediction using routine clinical data?},
  journal = {PLOS ONE},
  year    = {2017},
  volume  = {12},
  number  = {4},
  pages   = {e0174944},
  doi     = {10.1371/journal.pone.0174944}
}

@article{Gneiting2007,
  author  = {Gneiting, Tilmann and Raftery, Adrian E.},
  title   = {Strictly proper scoring rules, prediction, and estimation},
  journal = {Journal of the American Statistical Association},
  year    = {2007},
  volume  = {102},
  number  = {477},
  pages   = {359--378},
  doi     = {10.1198/016214506000001437}
}

@book{Cox1972,
  author    = {Cox, David R. and Oakes, David},
  title     = {Analysis of Survival Data},
  publisher = {Chapman and Hall},
  year      = {1984},
  address   = {London}
}

@book{Harrell2015,
  author    = {Harrell, Frank E.},
  title     = {Regression Modeling Strategies},
  publisher = {Springer},
  year      = {2015},
  edition   = {2},
  address   = {Cham}
}

@book{Rasmussen2006,
  author    = {Rasmussen, Carl Edward and Williams, Christopher K. I.},
  title     = {Gaussian Processes for Machine Learning},
  publisher = {MIT Press},
  year      = {2006},
  address   = {Cambridge, MA}
}

\end{document}